# A Self-organizing Interval Type-2 Fuzzy Neural Network for Multi-Step Time Series Prediction

Fulong Yao[a,*], Wanqing Zhao[a], Matthew Forshaw[a], Yang Song[b]

[a]*School of Computing, Newcastle University, Newcastle Upon Tyne, NE4 5TG, UK*

[b]*School of Mechanical and Electrical Engineering and Automation, Shanghai University, Shanghai, 200444, China*

**Abstract**

Data uncertainty is inherent in many real-world applications and poses significant challenges for accurate time series predictions. The interval type 2 fuzzy neural network (IT2FNN) has shown exceptional performance in uncertainty modelling for single-step prediction tasks. However, extending it for multi-step ahead predictions introduces further issues in uncertainty handling as well as model interpretability and accuracy. To address these issues, this paper proposes a new self-organizing interval type-2 fuzzy neural network with multiple outputs (SOIT2FNN-MO). Differing from the traditional six-layer IT2FNN, a nine-layer network architecture is developed. First, a new co-antecedent layer and a modified consequent layer are devised to improve the interpretability of the fuzzy model for multi-step time series prediction problems. Second, a new link layer is created to improve the accuracy by building temporal connections between multi-step predictions. Third, a new transformation layer is designed to address the problem of the vanishing rule strength caused by high-dimensional inputs. Furthermore, a two-stage, self-organizing learning mechanism is developed to automatically extract fuzzy rules from data and optimize network parameters. Experimental results on chaotic and microgrid prediction problems demonstrate that SOIT2FNN-MO outperforms state-of-the-art methods, by achieving a better accuracy ranging from 1.6% to 30% depending on the level of noises in data. Additionally, the proposed model is more interpretable, offering deeper insights into the prediction process.

## 1. INTRODUCTION

Time series data, a sequence of observations recorded at constant time intervals, is prevalent in various fields such as engineering [1, 2], economics [3, 4], meteorology [5] and health [6]. These data often exhibit temporal dependencies and trends, making them valuable for understanding and forecasting future system behaviours. However, time series observations in reality are always inherent to data uncertainties, such as stemming from noises in sensor measurements, disturbances in system operations and even errors in simulations and predictive models [7, 8]. Such uncertainties can propagate dramatically in the downstream modelling and decision-making processes. This brings fundamental challenges to time series prediction problems.

*Corresponding author: Fulong Yao, f.yao3@newcastle.ac.uk

Various machine learning methods, such as support vector machines (SVM) [9], long short-term memory (LSTM) networks [10] and convolutional neural network-LSTM (CNN-LSTM) [11], are commonly employed for time series forecasting. Specifically, LSTM networks are designed to address long-term dependencies by using a gating mechanism that enables them to retain information over extended time spans [12]. This allows LSTMs to capture complex temporal patterns, making them effective for sequential modelling. However, despite these advantages, LSTM-based models are often limited in capturing the diverse sources of uncertainty. This has led to suboptimal predictive performance and unreliable predictions [13], especially when further incorporated for decision making purposes.

In the past several decades, the fuzzy logic system (FLS) has been a great success in time series prediction, owing to the ability to handle linguistic and numerical uncertainties [14]. For example, Pourabdollah et al. [15] proposed a dynamic FLS based on a novel non-singleton fuzzification to improve the prediction accuracy of noisy time series (e.g., Mackey-Glass and Lorenz time series). Jafri et al. [16] developed a novel fuzzy logic for hourly wind prediction and achieved remarkable performance. However, FLS suffers from its root in handcrafted fuzzy rules and prior knowledge about the system, as well as potential challenges in generalizing to new data [17]. To address these limitations, researchers have explored a range of fuzzy neural networks (FNNs), such as adaptive-network-based fuzzy inference system (ANFIS) [18], combining the learning ability of artificial neural networks with the interpretability of fuzzy logic. These models can learn complex patterns from data, while incorporating fuzzy logic for semi-transparent reasoning. However, such models are designed based on type-1 fuzzy sets (with a crisp membership value), making it challenging to directly handle uncertainties in data [8, 19]. For example, in microgrid systems, renewable energy generation such as solar and wind power is highly uncertain and intermittent due to changing weather conditions [20]. Accurate short-term forecasting of these variables is crucial for operating the microgrid efficiently, enabling control actions such as importing additional electricity from the grid when the price is low and energy is clean. This requires more robust prediction approaches to ensure reliable energy management.

To remedy this issue, type-2 FNNs (T2FNNs) were developed, where type-2 fuzzy sets (T2FS) were employed to model uncertainties through the use of fuzzy (rather than crisp) membership functions [21]. This enhancement allows T2FNNs to process imprecise information more efficiently, without increasing the number of rules [22]. However, general T2FNNs often require extensive computation in type-reduction, which limits its real-world applications. This has led to the development of interval type-2 fuzzy neural networks (IT2FNNs), where interval type-2 fuzzy sets (IT2FS) are used in place of traditional T2FS, thereby significantly reducing computational complexity.

Recently, a number of IT2FNN approaches have been developed with a Takagi-Sugeno-Kang (TSK) consequent for handling data uncertainty and improving model accuracy. For example, Salimi-Badr [23] proposed an interval type-2 correlation-aware fuzzy neural network (IT2CAFNN) for nonlinear dynamic system modelling problems (including time-series predictions). To address the uncertainty, a shapeable IT2FS was designed to adaptively create the shape of fuzzy membership functions. Ashfahani et al. [24] developed an evolving type-2 quantum-based fuzzy neural network (eT2QFNN) for radio frequency identification (RFID) localization in a Manufacturing Shopfloor, in which an interval type-2 quantum fuzzy set with uncertain jump positions was designed to address noises in data. Apart from improving the antecedent of IT2FNNs, some studies were also conducted on enhancing the rule consequent. Beke and Kumbasar [25] designed a composite learning framework for interval type-2 fuzzy neural network (CLF-IT2NN), where a total of 12 CLF-IT2FNN forms were summarized based on combinations of different types of rule antecedents and consequents. Moreover, some researchers have tried to improve the performance of IT2FNN by modifying the network structure. For instance, Luo and Wang [26] proposed an interval type-2 LSTM fuzzy neural network (IT2FNN-LSTM) to improve the accuracy and uncertainty quantification for time series predictions. More related works can be found in [8, 27]. As a result of uncertainty modelling and an improved design of model architecture, the prediction accuracy can be improved. While an acceptable prediction error varies depending on the specific application, the literature has reported a range of errors, such as 16.56%-36.75 in Germany's power forecasting [28], and 5.34%-42.55% in Taiwan's photovoltaic generation prediction [29]. Moreover, recent studies [20, 30] have also examined how different levels (5%-25%) of prediction error could affect the control performance for operating a microgrid energy system.

Traditional IT2FNN models are consisted of the six-layer network (i.e., input layer, antecedent layer, rule layer, consequent layer, type reduction layer and output layer) [26], which may suffer from vanished rule firing strength when there is a large number of model inputs. Specifically, the increase in the number of model inputs will result in a decrease in the rule firing strength (due to fuzzy membership grade being between 0 and 1). When the input dimension is too large, all rules will fail to be fired for an effective model output, causing training to collapse. More discussions will be provided in Section 3.

As opposed to the single-step prediction problem, multi-step ahead predictions, though more informative in many applications (including economics [31], energy [32, 33] and aerology [34]), are found to be a very challenging task. At present, this can be done with three popular schemes: sliding window (SW), paralleling model (PM) and multiple outputs (MO) [35, 36, 37]. Generally, the SW scheme involves training a single-output model and then successively using this model to generate multiple predictions based on previously predicted values. However, this scheme relies on previously predicted values, which can accumulate prediction errors over time and lead to additional uncertainties. In contrast, the PM scheme involves training multiple single-output models, each making a prediction for a specific future time step. This obviously requires training and managing multiple predictive models, resulting in increased computational and storage requirements. Both of these schemes fall under single-step forecasting, which struggles to capture long-term dependencies in time series data [8]. As a result, this paper will investigate the MO scheme, where a multi-output model will be built, with each output giving a future prediction.

Although there are a few studies on IT2FNN for multi-step time series prediction using the MO scheme (e.g., [38] and [39]), they all ignored to model the temporal connections between multiple predictions, thereby reducing the prediction accuracy. Moreover, IT2FNN provides good interpretability in the single-output model through the design of IF-THEN rules, but this interpretability decreases significantly in the case of MO-based multi-step predictions, as each rule is required to account for a mixture of multiple predictions. Furthermore, IT2FNN can identify complex nonlinear system behaviours but at the expense of using a large number of fuzzy rules, which can lead to rule explosion [40]. Self-organizing IT2FNN with advanced rule learning capabilities has recently attracted widespread attention and proved promising in solving the rule explosion [41, 42, 43, 44]. However, current research in this field mainly focuses on single-step ahead predictions, with limited studies on multi-step predictions.

This paper aims to develop an efficient MO-based IT2FNN solution for multi-step time series prediction, focusing on two research questions: 1) How can the network architecture be designed to improve model accuracy, uncertainty handling and interpretability? 2) How can an effective learn-ing mechanism be developed to automatically extract the number of rules and model parameters from data? To address these questions, a new self-organizing interval type-2 fuzzy neural network with multiple outputs (SOIT2FNN-MO) is proposed, with the key contributions summarized as follows:

- a new link layer is proposed to improve accuracy by enhancing temporal connections between multi-step predictions;
- a new co-antecedent layer and a modified consequent layer are designed to improve the interpretability of fuzzy network;
- a new transformation layer is devised to address the potential issues in the vanishing rule firing strength caused by high-dimensional inputs;
- a two-stage self-organizing mechanism is developed to automatically generate the fuzzy rules, in which the first stage focuses on forming the rule base and performing initial optimization and the second stage is to determine all model parameters together.

The rest of this paper is structured as follows. Section 2 presents some preliminaries of IT2FNN and multi-step prediction. Section 3 and Section 4 present the network structure and learning mechanism of the proposed SOIT2FNN-MO, respectively. Section 5 provides a detailed performance evaluation, while Section 6 concludes the paper.

## 2. PRELIMINARIES

### *2.1. Interval Type 2 Fuzzy Set*

The concept of type-2 fuzzy sets (T2FS) was introduced by Zadeh [45] as an extension of an ordinary fuzzy set (type-1 fuzzy set (T1FS)) [46]. Unlike the T1FS whose membership grade is crisp, the degrees of membership in the T2FS are themselves fuzzy, also referred as 'fuzzy-fuzzy set' [47]. In this way, T2FS has the ability to deal with uncertain information that cannot be handled by a T1FS. However, T2FS-based systems exhibit computational complexity arising from the type 2 to type 1 reduction [48]. Given this, interval T2FS (IT2FS) was developed to strike a balance between computational efficiency and uncertainty handling.

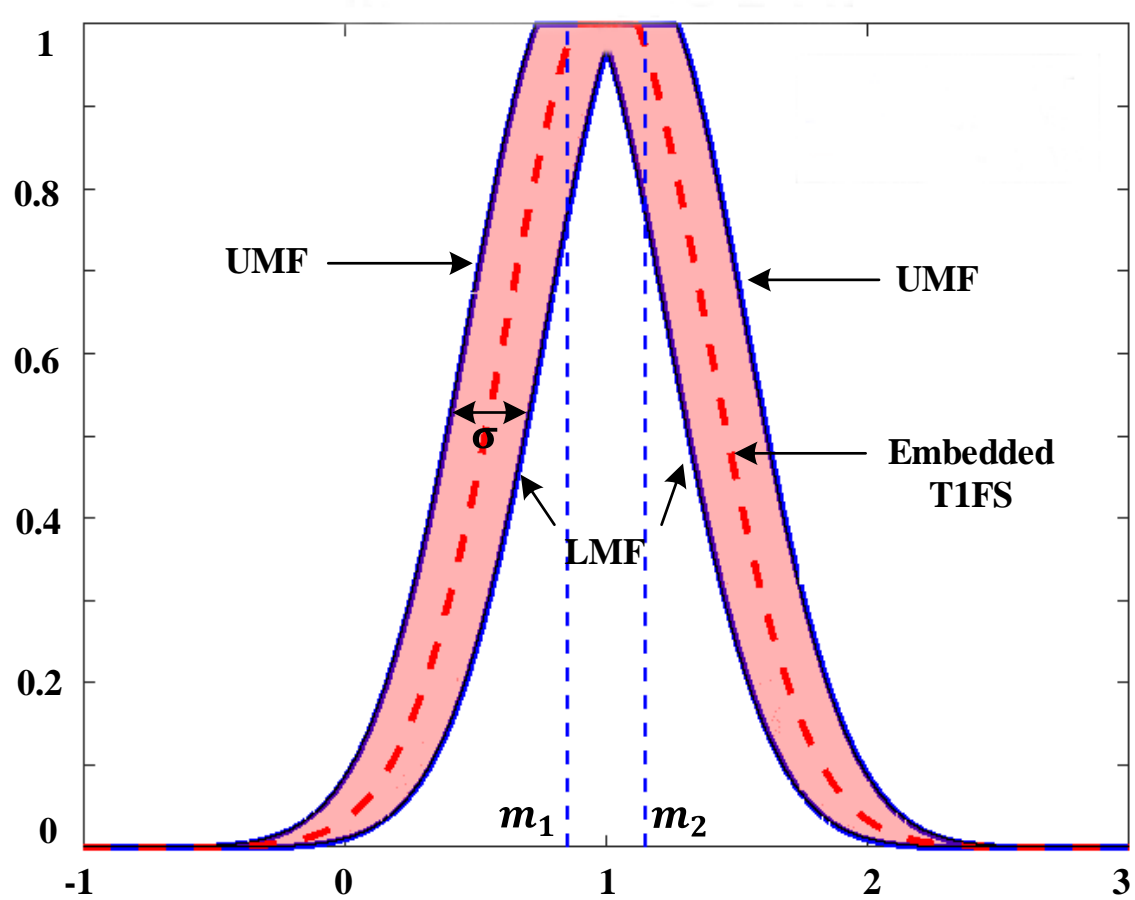

Figure 1: An IT2FS with an uncertain mean, adapted from [47]

The IT2FS is defined by an interval type 2 membership function (IT2MF) formed by an upper membership function (UMF) and a lower membership function (LMF). Fig. 1 illustrates an IT2FS with an uncertain mean. Here, uncertainty is modelled by representing membership values as intervals bound by the UMF and LMF [49]. This interval, known as the footprint of uncertainty (FOU), captures the range of possible membership degrees for each element, thereby enabling IT2FS to robustly handle uncertainties in data or linguistic expression. Actually, T1FS is a special case of IT2FS. As showed in Fig. 1, the dashed line represents an embedded T1FS, illustrating how determinism is embedded within uncertainty [47]. If the uncertainty (depicted by the pink shaded area) disappears, only the dashed line will exist. In other words, once the corresponding intervals becomes a constant subset, IT2FS will reduce to T1FS. Therefore, IT2FS gives additional degrees of uncertainty in designing fuzzy logic systems (such as IT2FNN).

### *2.2. IT2FNN Model*

An IT2FNN model is made up of an IT2FS antecedent and an interval set (or crisp set) consequent. At present, the IT2FNN with a Gaussian MF and a TSK (Takagi-Sugeno-Kang) consequent exhibits superiority over its peers, serving as a standard choice in many applications, such as weather [50] and energy [51]. The model can be expressed as a set of following IF-THEN rules:

$$
\begin{gathered}
Rule\ i : IF\ \ x_1\ is\ \tilde{A}_1^i\ AND ...\ AND\ x_n\ is\ \tilde{A}_n^i \\
THEN\ \ w^{i,k}\ \ is\ \ \tilde{a}_0^{i,k} + \sum_{j=1}^{n} \tilde{a}_j^{i,k} x_j
\end{gathered} \tag{1}
$$

where $x_1, ..., x_j, ..., x_n$ are model inputs, $\tilde{A}_j^i$ is the IT2FS of the *j-th* input with regard to the *i-th* rule, $w^{i,k}$ is the *k-th* output of the *i-th* rule, and $\tilde{a}_j^{i,k} = [c_j^{i,k} - s_j^{i,k}, c_j^{i,k} + s_j^{i,k}]$ is an interval set. In this way, the uncertainty in data can be captured by both antecedent and consequent sets.

### *2.3. Multi-step Prediction Schemes*

Multi-step prediction refers to the process of forecasting multiple future values in a time series. Unlike single-step prediction which forecasts the next immediate value only, multi-step prediction provides a more comprehensive outlook into the future. At present, there are three popular schemes: sliding window (SW), paralleling model (PM) and multiple outputs (MO).

*Sliding window (SW)*: This scheme involves iteratively forecasting future time steps using previously predicted values, forming a sequential prediction process [30]. For example, Fig. 2 (a) shows the prediction process of the SW scheme for a 3-step ahead prediction problem using the past 6 values. Specifically, a 6-input single-output model is built to always make the prediction for the next time step, and then this predicted value is successively used to compose the input for predicting the value of the following time step. As this 6-1 window continues to slide, three predicted outputs are generated in sequence.

*Paralleling model (PM):* This scheme involves the development of separate models, each making a single prediction for a different future time step [30]. Fig. 2 (b) gives the prediction process of the PM scheme. Instead of just training a model for generating all future predictions, it needs to build each single-output model individually which is costly and time consuming.

*Multiple outputs (MO):* This scheme uses a multi-output model to simultaneously predict multiple time series values into the future, without relying on the previous predictions [28]. Fig.

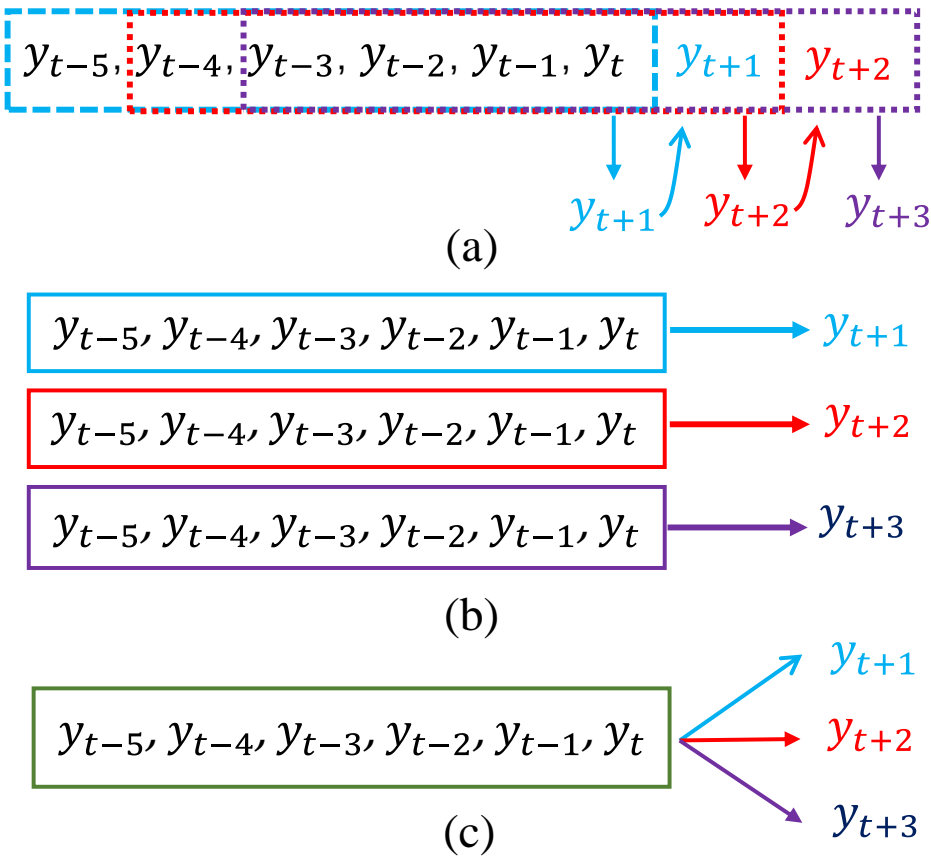

Figure 2: Multi-step prediction schemes

2 (c) illustrate the MO scheme. It trains a multi-output model capable of capturing complex temporal relationships across multiple time steps, providing all future predictions at one time.

## 3. SOIT2FNN-MO STRUCTURE

This section introduces an interval type-2 fuzzy neural network with multiple outputs (SOIT2FNN-MO) for multi-step time-series prediction problems. Fig. 3 shows the overall structure of the SOIT2FNN-MO. Differing from the traditional six-layer IT2FNN [52, 53, 54], a nine-layer network is devised here by introducing three additional layers (4, 5, 9) to improve the prediction accuracy and model interpretability, accounting for the nature of multi-step ahead predictions. In detail, a new co-antecedent layer (Layer 4) is designed to improve the interpretability of the rule antecedent for multiple outputs, a new transformation layer (Layer 5) to address the potential issues in the vanished rule firing strength, and a link layer (Layer 9) to enhance sequential connections among multiple predictions. In addition, modifications have been made to layer 6 (consequent layer) to enhance the interpretability of the rule consequent for predictions at different steps ahead. Each layer and its function are now detailed as follows:

*Layer 1 (Input Layer)*: The time-series input is given by $\tilde{x} = \{x_1, x_2, ...x_j, ..., x_n\}$, where $n$ is the total number of model inputs including scenarios of either univariate or multivariate inputs. For example, to predict from the past three values of 2 variables ($p_1$ and $p_2$), $\tilde{x}$ would be $\{p_1(t-2)$, $p_1(t-1)$, $p_1(t)$, $p_2(t-2)$, $p_2(t-1)$, $p_2(t)\}$. Each node in layer one represents a crisp input without any mathematical transformations.

*Layer 2 (Antecedent Layer)*: This layer is also referred to as the fuzzification layer. Here, each node employs a Gaussian interval type-2 membership function (IT2MF) to perform a fuzzification operation that converts crisp inputs into interval fuzzy values. In this paper, a Gaussian MF is employed with a fixed standard deviation but an uncertain mean value (see Fig. 1 ):

$$\mu_{\tilde{A}_j^i} = exp\{-\frac{1}{2}(\frac{x_j - m_j^i}{\sigma_j^i})^2\} = N(m_j^i, \sigma_j^i; x_j), m_j^i \in [m_{1,j}^i, m_{2,j}^i] \tag{2}$$

where $\tilde{A}_j^i$ is the IT2FS for the *j-th* input with regard to the *i-th* rule. Each Gaussian IT2MF can be further represented by an upper membership function (UMF) and a lower membership function

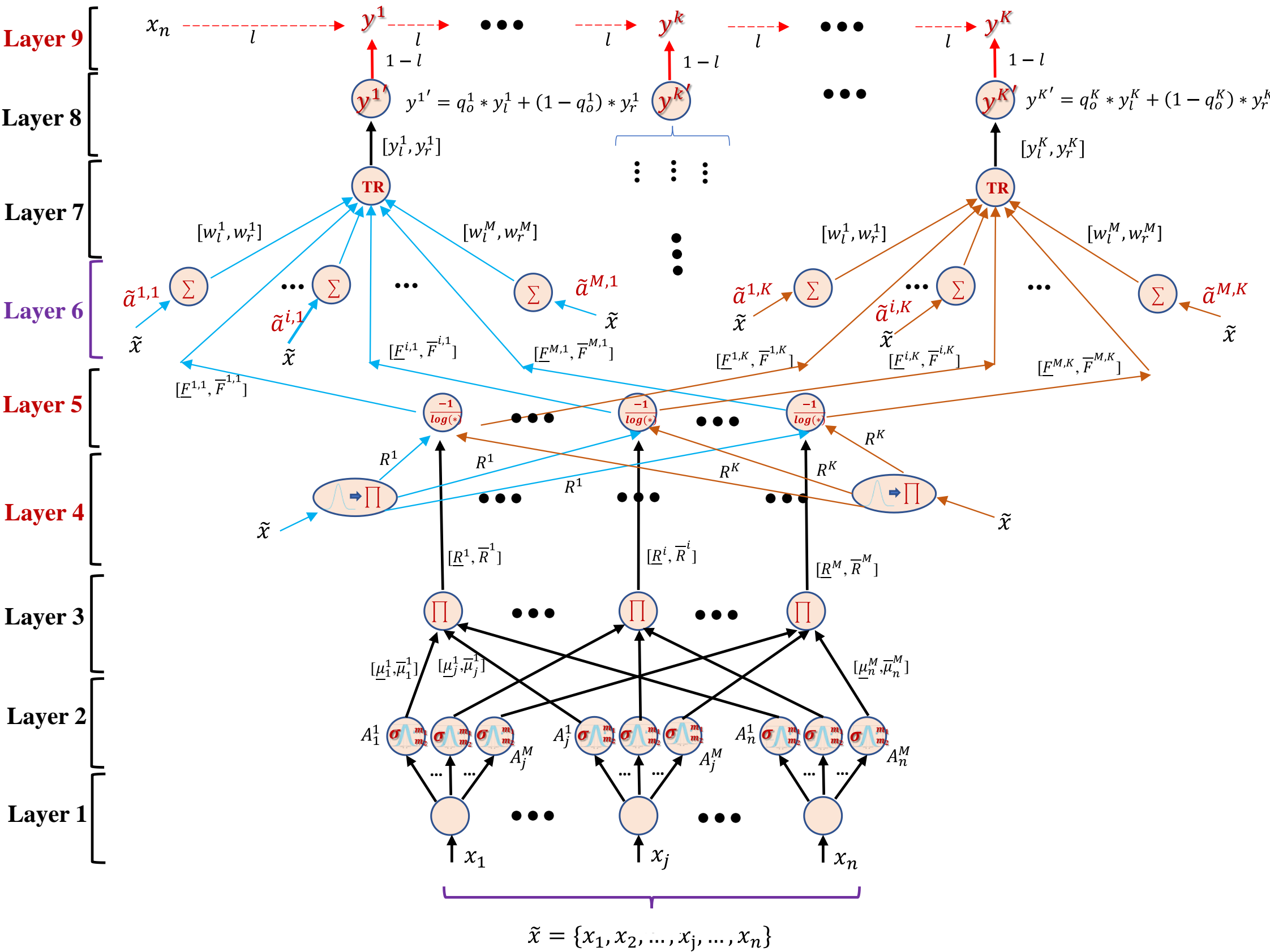

Figure 3: Structure of the proposed SOIT2FNN-MO

(LMF) as follows:

$$\overline{\mu}_j^i(x_j) = \begin{cases} N(m_{1,j}^i, \sigma_j^i; x_j), & if\ x_j < m_{1,j}^i; \\ 1, & if\ m_{1,j}^i \leq x_j \leq m_{2,j}^i; \\ N(m_{2,j}^i, \sigma_j^i; x_j), & if\ x_j > m_{2,j}^i. \end{cases} \tag{3}$$

$$\underline{\mu}_j^i(x_j) = \begin{cases} N(m_{2,j}^i, \sigma_j^i; x_j), & if\ x_j \leq \dfrac{m_{1,j}^i + m_{2,j}^i}{2}; \\ N(m_{1,j}^i, \sigma_j^i; x_j), & if\ x_j > \dfrac{m_{1,j}^i + m_{2,j}^i}{2}. \end{cases} \tag{4}$$

where $\sigma_j^i$ is the standard deviation of both LMF and UMF, and $m_{1,j}^i$ and $m_{2,j}^i$ are the means of LMF and UMF, respectively. Thus, the output of layer two is an interval $[\underline{\mu}_j^i(x_j), \overline{\mu}_j^i(x_j)]$. To ensure $\overline{\mu}_j^i(x_j) \geq \underline{\mu}_j^i(x_j)$ by the definition of IT2FS, the constraint $m_{1,j}^i \leq m_{2,j}^i$ should always be met. Here, data uncertainty is handled by using the Gaussian IT2FS, which represents an interval of membership values $[\underline{\mu}_j^i(x_j), \overline{\mu}_j^i(x_j)]$ rather than crisp points [44]. This enhances robustness and accuracy in prediction tasks when data is noisy or imprecise.

*Layer 3 (Rule Layer)*: This layer is also referred to as the firing strength of rules. This layer contains information regarding the influence range of each rule (there are $M$ rules in total, as shown in Fig. 3). Here, each node represents a fuzzy rule, and its output gives the lower ($\underline{R}^i$) and upper ($\overline{R}^i$) firing strength of this rule. Mathematically, the two firing strengths are computed by performing a fuzzy meet operation using the following algebraic product [52]:

$$\underline{R}^i = \prod_{j=1}^{n} \underline{\mu}_j^i(x_j); \quad \overline{R}^i = \prod_{j=1}^{n} \overline{\mu}_j^i(x_j) \tag{5}$$

*Layer 4 (Co-antecedent Layer)*: This layer is introduced to further improve the interpretability of the rule antecedent. There are $K$ nodes in this layer, each one corresponding to a specific time-series prediction (from 1-step ahead to K-step ahead). Within each node, two operations are performed, i.e., Gaussianization and algebraic product (6):

$$\mu_j^k(x_j) = N(m_j^k, \sigma_j^k; x_j); \quad R^k = \prod_{j=1}^{n} \mu_j^k(x_j) \tag{6}$$

where the index $k = 1, 2, ..., K$ represents the model outputs in sequence, as depicted in Layer 9 of Figure 3. Layer 4 is used to add an individual rule antecedent ($R^k$, in relation to the *k-th* model output) upon the existing one $[\underline{R}^i, \overline{R}^i]$, considering the distinct influence of an input vector $\tilde{x} = \{x_1, x_2, ...x_j, ..., x_n\}$. This layer provides an intuitive representation from the perspective of rule antecedent, i.e., how the input vector $\tilde{x}$ influences differently the prediction steps.

**Remark 1:** Although both the second and fourth layers incorporate Gaussian MFs, they differ in nature. The second layer employs the Gaussian IT2MF with an interval output, while the fourth layer adopts a normal Gaussian MF with a crisp output.

**Remark 2:** The second and fourth layers serve distinct purposes. The second layer functions as a shared layer across all model outputs (each IT2MF simultaneously affects all outputs), representing the commonalities across multiple model outputs. On the other hand, the fourth layer will only influence an individual model output, i.e., the prediction at a specific time step. By considering both the commonalities and individualities in the rule antecedent, the interpretability of the IT2FNN for time series predictions can thus be enhanced.

*Layer 5 (Transformation Layer)*: This layer is devised to address the potential issue of vanished rule firing strength that occurs in the case of high-dimensional inputs. Generally, in MO-based time series prediction, as the number of outputs increases, it is advisable to expand the input dimension to get more historical patterns and features [37]. However, since the Gaussian membership grade ($\mu_j^k(x_j)$, $\overline{\mu}_j^i(x_j)$ and $\underline{\mu}_j^i(x_j)$) is constrained to the range [0,1], an increased number of inputs can result in a rapid decline in the rule firing strength (see (5),(6)). In the model training process, this may cause numerical instability (generating values such as 'inf' or '-inf') and the problem of vanishing firing strength.

Here, by leveraging the properties of logarithmic operations, a novel aggregation function is devised to merge the shared firing strength interval $[\underline{R}^i, \overline{R}^i]$ with the individualized firing strength $R^k$ . This aggregation function employs the $log(.)$ to convert the product operation into a sum operation, effectively solving the vanished rule firing strength problem. Each node in this layer represents an aggregation function, producing an aggregated firing strength interval $F^{i,k}$ as the output. The expression for the aggregation function is expressed as:

$$F^{i,k} = [\underline{f}^{i,k}, \overline{f}^{i,k}], \quad i = 1, ..., M; \quad k = 1, ..., K \tag{7}$$

where $K$ is the number of model outputs (i.e., the number of steps to be predicted), and $\underline{f}^{i,k}$ and

$\overline{f}^{i,k}$ are defined as:

$$\underline{f}^{i,k} = -\frac{1}{\log(\underline{R}^i R^k)} = -\frac{1}{\sum_{j=1}^{n} log(\underline{\mu}_j^i(x_j)\mu_j^k(x_j))} = -\frac{1}{\sum_{j=1}^{n} log(\underline{\mu}_j^i(x_j)) + \sum_{j=1}^{n} log(\mu_j^k(x_j))} \tag{8}$$

$$\overline{f}^{i,k} = -\frac{1}{\log(\overline{R}^i R^k)} = -\frac{1}{\sum_{j=1}^{n} log(\overline{\mu}_j^i(x_j)\mu_j^k(x_j))} = -\frac{1}{\sum_{j=1}^{n} log(\overline{\mu}_j^i(x_j)) + \sum_{j=1}^{n} log(\mu_j^k(x_j))} \tag{9}$$

In this layer, $R^k$ is combined with all rules ($[\underline{R}^i, \overline{R}^i]$, i = 1,2,...,M) to produce $M$ aggregated intervals $[\underline{f}^{i,k}, \overline{f}^{i,k}]$. These intervals are then used together with the rule consequent from Layer 6 to generate the prediction for the $k$-$th$ model output. Therefore, each model output is related to all the rules and its corresponding $R^k$.

**Remark 3:** If the membership grades (e.g., $\underline{\mu}_j^i(x_j)$, $\overline{\mu}_j^i(x_j)$, and $\mu_j^k(x_j)$) lie in [0.0001, 1], the values of $log(x)$ will fall within the range of [-9.2103, 0]. Thus, the computation of $F$ is entirely manageable, thereby avoiding the issue of vanished rule firing strength. In fact, as long as the initial parameters of Gaussian MFs are appropriately chosen, occurrences of membership grades smaller than 0.001 are less likely.

**Remark 4:** $\underline{R}^i$ or $\overline{R}^i$ in Layer 3 is associated with a specific rule, whilst $R^k$ in Layer 4 is exclusively linked to a particular output. As a result, the interval $[\underline{f}^{i,k}, \overline{f}^{i,k}]$ is both rule-dependent and output-related. Moreover, the number of $R^k$ is solely determined by the number of outputs and is independent of the number of rules.

*Layer 6 (Consequent Layer)*: Each node in this layer represents a TSK rule consequent, operating as a linear combination of the model inputs $\tilde{x}$. Unlike traditional IT2FNN, where one rule (in layer 3) corresponds to just one TSK node, here each rule corresponds to $K$ nodes. Therefore, there are a total of $K * M$ nodes in this layer. The output of this layer can be represented as the following interval set $[w_l^{i,k}, w_r^{i,k}]$ in (10). In this way, the prediction at each step has a specific rule consequent, thus improving the interpretability of the consequent part of IT2FNN. Additionally, the interval set of this layer further enhances the uncertainty handling and reliability of the prediction network [25].

$$[w_l^{i,k}, w_r^{i,k}] = [c_0^{i,k} - s_0^{i,k}, c_0^{i,k} + s_0^{i,k}] + \sum_{j=1}^{n} [c_j^{i,k} - s_j^{i,k}, c_j^{i,k} + s_j^{i,k}] x_j; \tag{10}$$

The following equations can thus be derived:

$$w_l^{i,k} = \sum_{j=1}^{n} c_j^{i,k} x_j + c_0^{i,k} - \sum_{j=1}^{n} s_j^{i,k} |x_j| - s_0^{i,k}; \quad w_r^{i,k} = \sum_{j=1}^{n} c_j^{i,k} x_j + c_0^{i,k} + \sum_{j=1}^{n} s_j^{i,k} |x_j| + s_0^{i,k} \tag{11}$$

where $x_0 \equiv 1$, $w_*^{i,k}$ is the $k$-$th$ output of the $i$-$th$ rule. For $\forall i$ and $\forall k$, the consequent part must satisfy $w_l^{i,k} \leq w_r^{i,k}$; thus $s_0^{i,k} \geq 0$ and $s_j^{i,k} \geq 0$ hold [25].

*Layer 7 (Type Reduction Layer)*: This layer is responsible for converting type-2 into type-1 fuzzy sets. Each node in this layer corresponds to a linguistic output variable [55]. Instead of adopting traditional K-M iterative method [56], the output functions in (12) and (13) are used in this paper to perform the type reduction more efficiently. Here, the factors $q_l^k$ and $q_r^k$ are employed to adaptively adjust the lower and upper positions of the *k-th* interval output $[y_l^k, y_r^k]$.

$$y_l^k = \frac{(1-q_l^k)\sum_{i=1}^{M} \underline{f}^{i,k} w_l^{i,k} + q_l^k \sum_{i=1}^{M} \overline{f}^{i,k} w_l^{i,k}}{\sum_{i=1}^{M}(\underline{f}^{i,k} + \overline{f}^{i,k})} \tag{12}$$

$$y_r^k = \frac{(1-q_r^k)\sum_{i=1}^{M} \underline{f}^{i,k} w_r^{i,k} + q_r^k \sum_{i=1}^{M} \overline{f}^{i,k} w_r^{i,k}}{\sum_{i=1}^{M}(\underline{f}^{i,k} + \overline{f}^{i,k})} \tag{13}$$

Noted, in the above equations, $q_l^k \in [0,1]$ and $q_r^k \in [0,1]$

*Layer 8 (Defuzzification Layer)*: Each node in this layer gives a crisp output that corresponds to the prediction at a future time step. This can be computed by the following defuzzification equation:

$$y^{k'} = q_o^k y_l^k + (1-q_o^k) y_r^k \tag{14}$$

where $q_o^k \in [0,1]$ is the weight that balances the importance between $y_l^k$ and $y_r^k$ [8].

*Layer 9 (Link Layer)*: In MO-based time series prediction, it usually assumes that the model outputs are independent of each other, which breaks the temporal connections between predictions at different time steps. To make the proposed SOIT2FNN-MO suitable for multi-step ahead predictions, a new link layer is designed here to build such a connection, as expressed in (15). This layer ensures that the predictions are not only determined by the model inputs, but also affected by predictions from its preceding steps. This will ultimately improve the prediction accuracy as well as its stability.

$$y^k = \begin{cases} (1-l)y^{k'} + l x_n, & if\ k=1; \\ (1-l)y^{k'} + l y^{k-1}, & if\ k>1. \end{cases} \tag{15}$$

Here, $l \in [0,1]$ is the weight factor balancing the impact between the current and preceding predictions.

## 4. LEARNING METHOD

Given the proposed SOIT2FNN-MO, this section presents a two-stage self-organizing learning mechanism to determine both the model structure and parameters. The implementation steps are summarised in Algorithm 1. Here, the first stage is used to create the rule base from empty and perform initial parameter optimization, while the second stage is designed to fine-tune all model parameters together. It should be noted that the status flag $F_s$ can take three possible values: 0 indicates that Stage 1 is running (i.e., in the rule growing or rule removing steps), 1 indicates that Stage 1 has ended (i.e., there is no increase or decrease in rules in the current episode) and 2 indicates the completion of Stage 2 (i.e., a global optimization has been completed).

### *4.1. Pre-stage: Normalization and Clustering*

To prevent the distribution of membership grades from becoming overly dispersed, normalizing the inputs is an important step. In this paper, the max-min scaling is simply employed to perform the normalization. Then, Fuzzy C-Means (FCM) [57] is employed in this paper to generate an initial number of $N_c$ clusters, which will serve as the basis for structure learning of SOIT2FNN-MO.

Partitioning the input space into local regions with similar system behaviours through clustering can effectively enhance the adaptability of IT2FNN model to diverse data patterns [53, 58]. The derived clusters can be used to build the rule base with reduced computational complexity and improved model interpretability [59]. Here, the centroid and width of each cluster has the opportunity to be selected to initialize antecedent parameters in layer 2, thus generating a potential fuzzy rule. Once a cluster is selected, antecedent parameters can be determined as:

$$m_{1,j}^{i} = m_j^{\mathfrak{c}}(1-\Upsilon); \quad m_{2,j}^{i} = m_j^{\mathfrak{c}}(1+\Upsilon); \quad \sigma_j^{i} = \sigma_j^{\mathfrak{c}} \tag{16}$$

where $\Upsilon = 0.1$ represents the uncertainty in the mean value (centroid) of the cluster, $m_j^{\mathfrak{c}}$ and $\sigma_j^{\mathfrak{c}}$ are the mean and STD of $\mathfrak{c}$-$th$ cluster, respectively.

The initialization of co-antecedent parameters in layer 4 is much easier. Each co-antecedent MF corresponds to one output, and the quantity does not vary with the total number of rules. Given this, the initial co-antecedent parameters (e.g., $m_j^k$ and $\sigma_j^k$) can be directly set to the mean and standard deviation of the set composed of all $x_j$, applicable to all co-antecedent MFs:

$$m_j^k = m_j^c; \quad \sigma_j^k = \sigma_j^c \tag{17}$$

where $m_j^c$ and $\sigma_j^c$ are the mean and standard deviation of the $j$-$th$ input set.

### *4.2. Stage 1: Self-organizing Structure Learning*

#### *4.2.1. Rule Growing Step*

The rule growing step is devised to generate rules in sequence (i.e., each episode can generate at most one rule) until a new rule no longer brings a significant improvement in the prediction accuracy. For each potential rule generation, the flowchart is shown in Fig. 4 and the implementations are depicted in Algorithm 2.

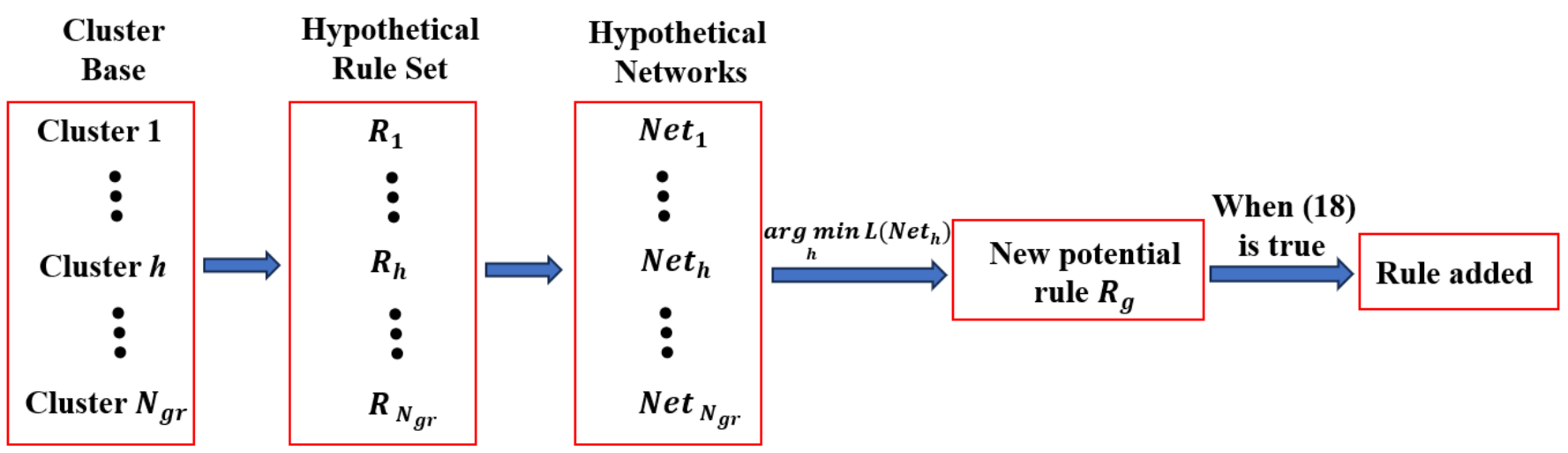

Figure 4: Flowchart of the rule growing step

Specifically, each cluster obtained from FCM is used to make up a new hypothetical rule $R_h$. This hypothetical rule, together with the existing rules, forms a hypothetical network $Net_h$. In this hypothetical network, the antecedent parameters (in Layer 2) are directly determined by the selected clusters as in (16). The remaining parameters (a total number of $2 \cdot K \cdot n + (n+1) \cdot 2 \cdot M \cdot K + 2 \cdot K + K + 1$ in Layers 4,6,7,8,9) are optimized using nonlinear least square (*lsqcurvefit* in Matlab). In this way, $N_{gr}$ hypothetical rules will be generated here, but only the hypothetical rule corresponding to the network $Net_h$ with the smallest loss $L_g$ (defined as mean squared error) is qualified to become a new potential growing rule $R_g$. However, this potential rule is only going to be added to the existing model if it leads to a significant loss reduction:

$$L_e - L_g \geq T_g \tag{18}$$

**Algorithm 1:** Learning Method of SOIT2FNN-MO

**Input** : $\tilde{x}, \tilde{y}$

**Procedure:**

1 Normalize $\tilde{x}$ and initialize learning parameters and model weights;
2 Group $\tilde{x}$ into $N_c$ clusters using $FCM$ and store them in a cluster base $B^C$;
3 Define the rule number $N_r = 0$, status flag $F_s = 0$;
4 Create an empty rule base $B^R$ and an empty base $B^S$ to store generated rules and selected clusters, respectively;
5 **for** $i$ = *1 to* $Episode_{max}$ **do**
6   - - - - - - - - - - - - - **Stage 1** - - - - - - - - - - - - -
7   Jump to **Algorithm 2** for identifying the next potential rule to be added;
8   **if** $L_e - L_g \geq T_g$ **then**
9     Add the new rule $R_g$ into rule base $B^R$;
10     Assign $L_e = L_g$; $F_s = 0$;
11     Move $C_g$ from $B^C$ to $B^S$;
12     Update parameters using $Net_g$;
13   **else**
14     **if** $Size(B^R) \leq 1$ **then**
15       **if** $F_s == 2$ **then**
16         **break**;
17       **else**
18         $F_s = 1$;
19     **else**
20       Jump to **Algorithm 3** for identifying the least significant rule to be removed;
21       **if** $L_r - L_e < T_r$ **then**
22         **if** $F_s == 2$ **then**
23           **break**;
24         **else**
25           $F_s = 1$;
26       **else**
27         Remove $R_r$ from the rule base $B^R$;
28         $L_e = L_r$; $F_s = 0$;
29         Move $C_r$ from $B^S$ back to $B^C$;
30         Update parameters using $Net_r$;
31   - - - - - - - - - - - - - **Stage 2** - - - - - - - - - - - - -
32   **if** $F_s$ == *1* **then**
33     Jump to **Algorithm 4** for global optimization;
34     $L_e = L_{gl}$, then update parameters using $Net_{gl}$;
35     $F_s = 2$;

**Output** : Trained SOIT2FNN-MO

**Algorithm 2:** Rule Growing Step

**Input** : $\tilde{x}, \tilde{y}$, $B^C$

**Procedure:**

1 $N_{gr}$ = Size($B^C$);
2 Create an empty array $L$ to store loss;
3 **for** $h$ = *1 to* $N_{gr}$ **do**
4 Pre-generate a new hypothetical rule $R_h$;
5 Randomly initialize the consequent parameters $c_j^{i,k}$, $s_j^{i,k}$ of $R_h$;
6 Initialize the weight parameters of $R_h$ as $q_l^k = q_r^k = q_o^k = 0.5$ ;
7 Initialize $R_h$'s antecedent IT2MF and co-antecedent MF using using (16) - (17) ;
8 Fix antecedent parameters and optimize all other parameters over the network $Net_h$ using *lsqcurvefit*;
9 Compute the loss $L_h$ of the whole training set;
10 Concatenate $L_h$ to the end of $L$ ;
11 Sort loss $L$, then find the minimum one $L_g$ and its corresponding rule $R_g$, cluster $C_g$ and Net $Net_g$.

**Output** : $L_g$, $R_g$, $C_g$, $Net_g$

**Algorithm 3:** Rule Removing Step

**Input** : $\tilde{x}, \tilde{y}$, $B^R$, $B^S$

**Procedure:**

1 $N_{re}$ = Size($B^R$);
2 Create an empty array $L$ to store loss;
3 **for** $h$ = *1 to* $N_{re}$ **do**
4 Remove the *h-th* hypothetical rule from $B^R$ and denote the network composed of remaining rules as $Net_h$;
5 Fix antecedent parameters and optimize other parameters of $Net_h$ using *lsqcurvefit*;
6 Use the $Net_h$ to compute the loss $L_h$ over the whole training set;
7 Concatenate $L_h$ to the end of $L$ ;
8 Sort loss $L$, then find the minimum one $L_r$ and its corresponding rule $R_r$ and cluster $C_r$ and $Net_r$.

**Output** : $L_r$, $R_r$, $C_r$, $Net_r$

**Algorithm 4:** Global Network Optimization

**Input** : $\tilde{x}, \tilde{y}$, $B^R$

**Procedure:**

1 Optimize all parameters of $Net_{gl}$ together using SGD, i.e., (21) ;
2 Compute the loss $L_{gl}$ of the whole training set;

**Output** : $Net_{gl}$

where $T_g$ is the threshold for adding a new rule, and $L_e$ is the loss value of the existing network. To ensure a smooth growth of rules, the initial value of $L_e$ is typically set to a very large value. Here, it is configured as 1e10. It should be noted that, in the rule growing process, the common parameters in Layer 2 are kept unchanged, while all individual parameters in other layers are optimized together.

*4.2.2. Rule Removing Step*

Once the rules stop growing, the rule removing step is introduced to prune insignificant rules for a compact network structure. This is carried out by excluding rules from the current network in sequence (each episode can remove at most one rule), under the condition that the removal of a rule will result in a negligible increase in the loss value. The actual process is depicted in Algorithm 3. Here, each rule in the current network becomes a hypothetical removing rule $R_h$. The remaining rules after excluding $R_h$ then construct a new hypothetical network $Net_h$. Similar to the rule growing step, each hypothetical network is optimized under fixed antecedent parameters. The hypothetical rule corresponding to the network $Net_h$ with the smallest loss $L_r$ becomes the potential rule $R_r$ to be removed.

Then, only if the loss increase caused by this potential rule pruning falls below a certain threshold (see (19)), it can be removed from the current network. Otherwise, the rule is considered important without removal and the algorithm moves on to Stage 2.

$$L_r - L_e < T_r \tag{19}$$

Here, $T_r$ is the threshold for removing an existing rule. Furthermore, to ensure that the algorithm does not get stuck in an endless loop, $T_r \leq T_g$ must hold.

*4.3. Stage 2: Parameter Tuning*

In stage 1, at most one rule is allowed to be added or removed per episode. When the number of rules changes, the network undergoes a local parameter optimization. However, if the rule number remains constant throughout an episode, stage 2 is initiated to tune the existing parameters globally. The implementation of stage 2 is depicted in Algorithm 4. Differing from stage 1, all parameters here $(n \cdot 3 \cdot M + 2 \cdot K \cdot n + (n+1) \cdot 2 \cdot M \cdot K + 2 \cdot K + K + 1)$ are optimized using SGD. The loss function $E$ can be represented as:

$$E = \frac{1}{2} \sum_{k=1}^{K} (y^k - y_a^k)^2 \tag{20}$$

where $y_a^k$ is the actual value of the $k$-$th$ output.

Then, the parameters can be updated by gradient descent:

$$V = V - \eta \frac{\partial E}{\partial V} \tag{21}$$

where $\eta$ is the learning rate and $V$ represents the model parameters, i.e., $c_j^{i,k}$, $s_j^{i,k}$, $m_{1,j}^i$, $m_{2,j}^i$, $\sigma_j^i$, $m_j^k$, $\sigma_j^k$, $q_l^k$, $q_r^k$, $q_o^k$, $l$. The derivatives of the loss function with respect to these variables can be found in Section 1 of the supplementary materials.

In summary, the learning process begins with a pre-stage dedicated to identifying the pool of initial fuzzy rules using FCM clustering. The training process then alternates between two stages. In stage 1, the model structure is updated in two steps by adding or removing a rule into the model. This gives the updated rule base and locally optimized parameters (excluding the fixed antecedent parameters). Then, stage 2 is used to fine-tune all parameters, where all network parameters are globally optimized. The training continues cycling through these two stages until the rule base no longer changes, at which point the final SOIT2FNN-MO model is obtained. It is important to note that even if no rules are added or removed in stage 1, the optimization in stage 2 may still give a new possibility for rule change in the subsequent iterations. Therefore, the final model structure is confirmed only when the number of rules remains unchanged after one complete round of stage 2 and stage 1 in sequence.

## 5. SIMULATION AND RESULTS

This section presents a comprehensive evaluation of the proposed approach on both simulated and real-world datasets. First, SOIT2FNN-MO is compared with other state-of-the-art approaches for time series forecasting, demonstrating its superiority in the prediction accuracy and resilience to uncertainty. This is followed by a detailed analysis regarding the structural design and sensitivity testing (i.e., cluster numbers). The root mean square error (RMSE) and mean percentage error (MPE) are employed as the evaluation metrics. All simulations were conducted in MATLAB under Windows 10 operating system, Intel Core i7-1185G7 3.00GHz 32.0 GB.

### *5.1. Example 1 (Chaotic Time Series Prediction)*

The performance of SOIT2FNN-MO is first evaluated on the Mackey–Glass chaotic time series. The data was generated using the following delay differential equation (DDE):

$$\frac{dx(t)}{dt} = \frac{0.2x(t-\tau)}{1+x^{10}(t-\tau)} - 0.1x(t) \tag{22}$$

where $\tau \geq 17$. As depicted in [60, 61], the system response was chaotic time series. Here, the initial conditions were set as: $\tau = 30$ and $x(0) = 1.2$ according to [52, 62, 63].

A total of 1500 data points were generated from the interval $t \in [31, 1530]$ (see Fig. 5). Here, the first 1000 points were employed for training while the remaining 500 points were used for testing. Then, a 9-input and 3-output prediction problem was formulated as $[x(t-16), x(t-14), x(t-12), x(t-10), x(t-8), x(t-6), x(t-4), x(t-2), x(t); x(t+2), x(t+4), x(t+6)]$. The relevant parameters were set as: $T_g = T_r = 0.0025$, $\eta = 0.03$, $N_c = 5$ and $Episode_{max} = 100$. In addition, the number of iterations in the optimization within the steps of rule growing/removing and parameter fine-tuning were set as 1000 and 3000, respectively.

Moreover, a composite learning framework for interval type-2 fuzzy neural network (CLF-IT2NN) [25] and a machine learning approach (CNN-LSTM [11]) were employed for comparisons. It should be noted that Beke and Kumbasar [25] listed a total of 12 types of CLF-IT2NN based on different rule antecedents and consequents. Here, S-IVL was chosen to ensure consistency with the proposed approach (i.e., Gaussian antecedent and TSK consequent). Additionally, to avoid the problem of vanished rule firing strength, a transformation layer ($log(.)$) was incorporated into CLF-IT2NN.

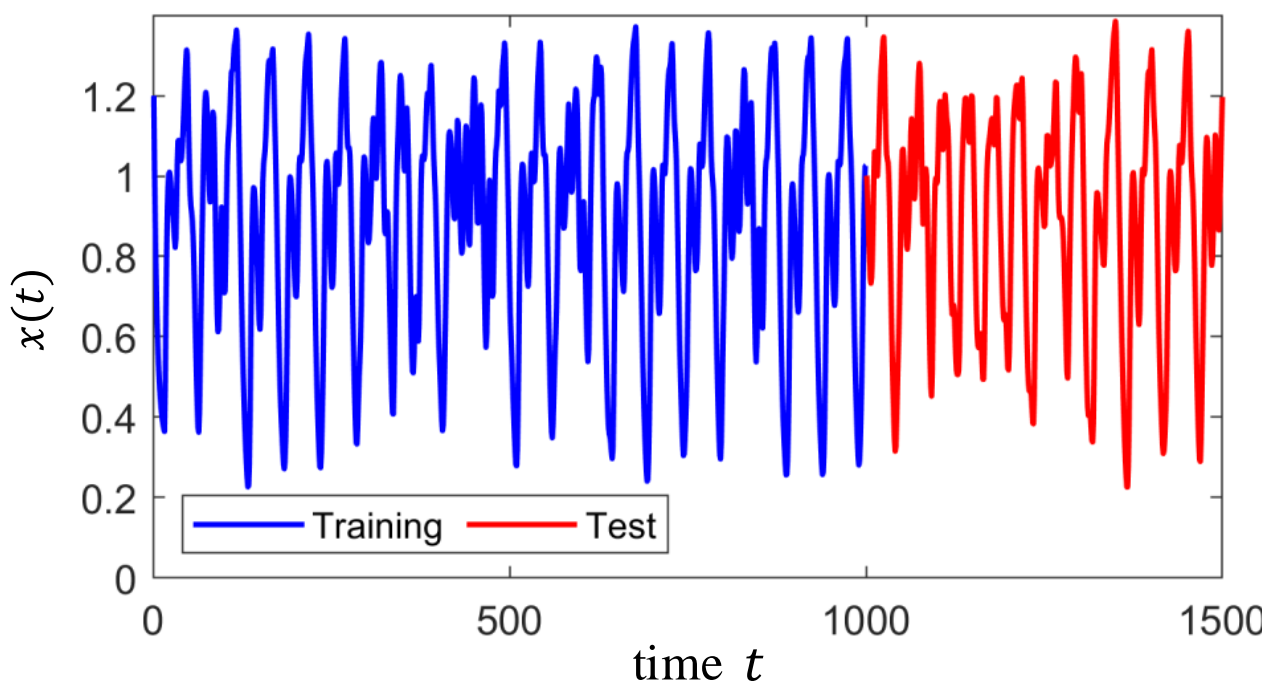

Figure 5: An illustration of Chaotic time series

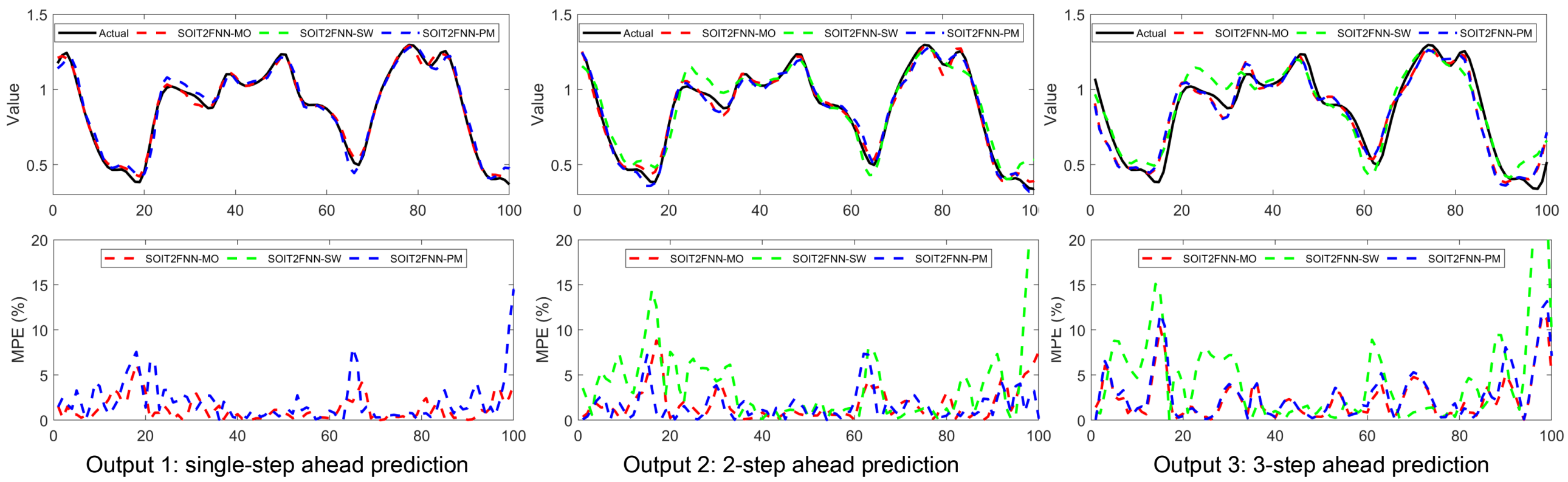

Figure 6: A comparison of the prediction results on clean (test) chaotic time series

Apart from the noise-free conditions, the performance of these algorithms is also evaluated against different levels of noises in the data (i.e., adding Gaussian noise with STDs of 10% and 30%, respectively). Fig. 6 shows an illustration of the predicted and actual values (over the test dataset without noises) using SOIT2FNN, based on SW, PM and MO schemes, respectively. Moreover, Tables 1 - 3 present the performance comparisons on both clean and noisy time series, where the results are averages of the three predictions. A more detailed comparison of RMSE and MPE for each time-step prediction (i.e., $y^1 - y^3$) was provided in Tables S1-S3 in supplementary materials.

It can be observed that the proposed SOIT2FNN-MO models generally performed better than CLF-IT2FNN (both in RMSE and MPE), especially for the noisy data. Despite that CNN-LSTM achieved better results than SOIT2FNN-MO on the clean dataset, its performance rapidly declined with the increase of noise. In particular, the proposed approach far outperformed CNN-LSTM when getting added noises with $std = 30\%$, demonstrating exceptional robustness to uncertainty. When models trained with noisy data (i.e., $std = 10\%$ and $std = 30\%$), the test performance of CNN-LSTM dropped to an unacceptable level. Another interesting finding concerns the effectiveness of the three schemes for multi-step ahead predictions. MO and PM exhibited greater resilience to noises in data compared to SW. This was attributed to the cumulative error gained in applying the sliding window [64]. Similar conclusions were reported in previous studies [37] and [65], where MO and PM schemes were seen more robust than SW in various applications. More-over, although PM performed similarly to MO on clean and low-noise (10%) datasets, it became much worse than the latter when the noise increased to 30%.

Table 1: Performance comparison on models trained with clean chaotic time series

| Data | CNN-LSTM [11] | | | CLF-IT2NN [25] | | | **SOIT2FNN** | | |
|---|---|---|---|---|---|---|---|---|---|
| | SW | PM | MO | SW | PM | MO | SW | PM | **MO** |
| Training RMSE (Clean) | 0.01 | 0.01 | 0.01 | 0.04 | 0.04 | 0.04 | 0.06 | 0.04 | **0.03** |
| Test RMSE (Clean) | 0.01 | 0.01 | 0.01 | 0.04 | 0.04 | 0.04 | 0.06 | 0.04 | **0.03** |
| Test RMSE (std=10%) | 0.15 | 0.15 | 0.15 | 0.20 | 0.17 | 0.17 | 0.19 | 0.18 | **0.17** |
| Test RMSE (std=30%) | 0.39 | 0.39 | 0.39 | 0.62 | 0.53 | 0.53 | 0.81 | 0.49 | **0.50** |
| Training MPE (Clean) | 0.90 | 0.86 | 0.72 | 4.89 | 3.13 | 3.30 | 5.41 | 3.45 | **2.71** |
| Test MPE (Clean) | 0.92 | 0.88 | 0.77 | 3.80 | 3.71 | 3.77 | 5.67 | 3.69 | **3.16** |
| Test MPE (std=10%) | 16.2 | 15.9 | 16.2 | 21.4 | 19.1 | 19.0 | 19.9 | 18.8 | **18.6** |
| Test MPE (std=30%) | 77.4 | 75.0 | 74.9 | 70.3 | 68.3 | 68.4 | 86.9 | 67.5 | **67.2** |
| Rule | — | — | — | 2 | 2,2,2 | 2 | 1 | 1,2,2 | **2** |

Table 2: Performance comparison on models trained with noisy chaotic time series (std = 10%)

| Data | CNN-LSTM [11] | | | CLF-IT2NN [25] | | | **SOIT2FNN** | | |
|---|---|---|---|---|---|---|---|---|---|
| | SW | PM | MO | SW | PM | MO | SW | PM | **MO** |
| Training RMSE (std=10%) | 0.14 | 0.12 | 0.12 | 0.13 | 0.13 | 0.13 | 0.13 | 0.13 | **0.13** |
| Test RMSE (Clean) | 0.06 | 0.05 | 0.05 | 0.06 | 0.06 | 0.05 | 0.05 | 0.05 | **0.05** |
| Test RMSE (std=10%) | 0.14 | 0.14 | 0.14 | 0.19 | 0.17 | 0.15 | 0.14 | 0.14 | **0.14** |
| Test RMSE (std=30%) | 0.38 | 0.36 | 0.36 | 0.40 | 0.39 | 0.37 | 0.35 | 0.35 | **0.35** |
| Training MPE (std=10%) | 14.5 | 14.4 | 14.2 | 15.0 | 14.0 | 15.2 | 14.9 | 14.7 | **14.8** |
| Test MPE (Clean) | 5.68 | 5.00 | 5.12 | 5.55 | 5.22 | 5.22 | 5.39 | 5.09 | **4.90** |
| Test MPE (std=10%) | 14.6 | 15.9 | 15.9 | 18.4 | 16.6 | 16.3 | 15.9 | 15.5 | **15.4** |
| Test MPE (std=30%) | 83.8 | 81.4 | 76.1 | 51.2 | 46.1 | 45.2 | 41.5 | 41.2 | **41.4** |
| Rule | — | — | — | 2 | 2,2,3 | 3 | 2 | 2,2,2 | **2** |

### *5.2. Example 2 (Microgrid Monitoring)*

Multi-step time series forecasting in a microgrid system presents as an essential task for energy network monitoring and control. This can involve the prediction of electricity price and import/export energy (i.e., unmet power due to the lack/surplus of on-site renewable supplies) to meet the power demand of a locality at the lowest cost. In this example, the proposed SOIT2FNN-MO model is evaluated on a real-world time series dataset [66, 20] collected from a US district microgrid system. This dataset contains unmet power and electricity price at a hourly resolution over a year. Here, to cover the seasonality effect, sensor measurements from the first 21 days of each month were extracted to form the training set, while the remaining data was used as test set. As a result, there are 6048 points in the training set and 2736 points in the test set, as shown in Fig. 7.

Here, the time series of the past nine time steps (hours) was used to predict the values at the next three time steps. Unlike an autoregressive prediction (i.e., chaotic time series prediction), three additional variables were incorporated into the model inputs to improve prediction accuracy and stability. Specifically, the time features (i.e., month, weekday and hour) of the current time step of each time series were used. Thus, the proposed SOIT2FNN-MO model has 12 inputs and

Table 3: Performance comparison on models trained with noisy chaotic time series (std=30%)

| Data | CNN-LSTM [11] | | | CLF-IT2NN [25] | | | **SOIT2FNN** | | |
|---|---|---|---|---|---|---|---|---|---|
| | SW | PM | MO | SW | PM | MO | SW | PM | **MO** |
| Training RMSE (std=30%) | 0.11 | 0.07 | 0.11 | 0.37 | 0.40 | 0.36 | 0.36 | 0.42 | **0.34** |
| Test RMSE (Clean) | 0.34 | 0.31 | 0.32 | 0.19 | 0.21 | 0.18 | 0.17 | 0.24 | **0.16** |
| Test RMSE (std=10%) | 0.38 | 0.34 | 0.34 | 0.26 | 0.23 | 0.19 | 0.21 | 0.27 | **0.19** |
| Test RMSE (std=30%) | 0.64 | 0.59 | 0.56 | 0.41 | 0.44 | 0.39 | 0.36 | 0.41 | **0.36** |
| Training MPE (std=30%) | 16.3 | 9.27 | 16.3 | 30.0 | 30.4 | 26.3 | 29.7 | 34.9 | **25.9** |
| Test MPE (Clean) | 32.8 | 31.4 | 32.1 | 18.9 | 19.7 | 16.7 | 21.5 | 22.8 | **15.7** |
| Test MPE (std=10%) | 38.4 | 36.6 | 35.1 | 26.3 | 23.5 | 23.0 | 27.4 | 27.1 | **22.0** |
| Test MPE (std=30%) | 115 | 102 | 96.1 | 57.2 | 59.3 | 48.6 | 55.0 | 46.2 | **44.5** |
| Rule | — | — | — | 3 | 2,3,3 | 5 | 2 | 2,3,3 | **5** |

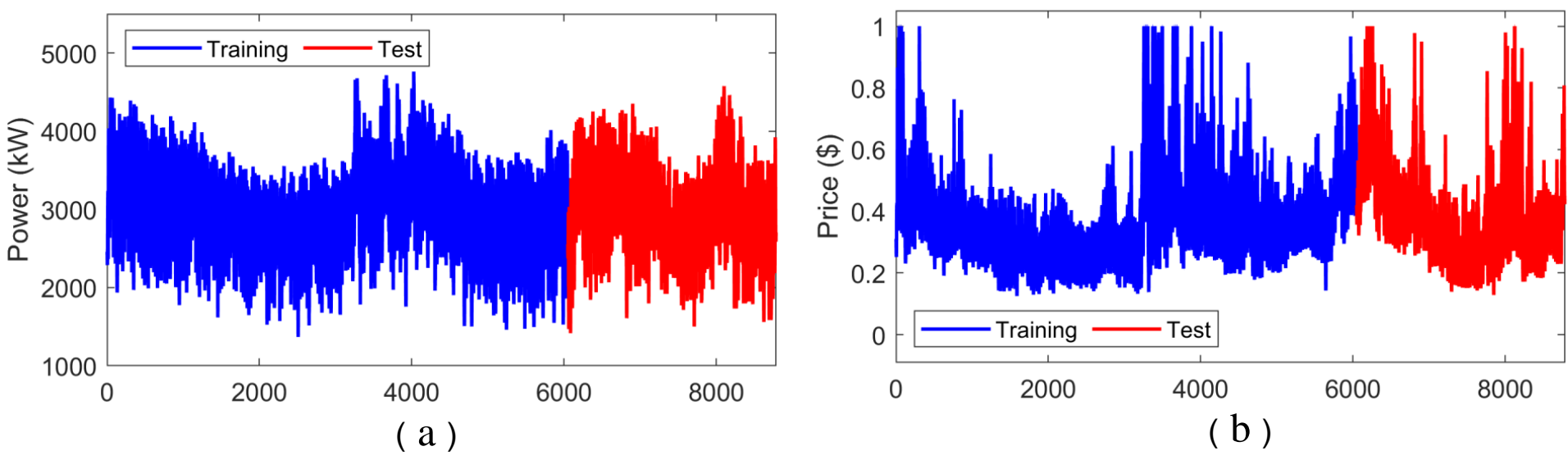

Figure 7: Illustrations of the unmet power (a) and electricity price (b)

3 outputs, denoted as: $[m(t), w(t), h(t), v(t-8), v(t-7), v(t-6), v(t-5), v(t-4), v(t-3), v(t-2), v(t-1), v(t); v(t+1), v(t+2), v(t+3)]$. Here, $v(t)$ is the unmet power or electricity price at time instant $t$, while $m(t) \in [1, 12]$, $w(t) \in [1, 7]$, $h(t) \in [0, 23]$ are the month, weekday and hour features at current time instant. Similar as in Example 1, Gaussian noises with variances of 10% and 30% were added to the dataset to simulate data uncertainty within the microgrid system. These parameters were set in the model determination process: $N_c = 5$ for unmet power prediction, $N_c = 10$ for price prediction and $T_g = T_r = 0.001$ for both predictions. Other parameters were kept the same as those used in the previous example.

Figs. 8 and 9 show the predicted and actual values for the unmet power and electricity price (from 00:00 25th Jan to 03:00 29th Jan). Tables 4 and 5 present the performance comparisons on clean and noisy datasets. More details on each time step predictions were listed in Tables S4 and S5 of the supplementary materials. It is evident that SOIT2FNN-MO still outperformed CLF-IT2FNN in this real-world dataset, even in the presence of added noises. Compared to Chaotic time series, the proposed approach demonstrated better unceratinty handling in this example. For instance, SOIT2FNN-MO significantly outperformed CNN-LSTM in unmet power prediction when noise level was set to $std = 10\%$. This occurred earlier compared to the chaotic time series prediction, where SOIT2FNN-MO outperformed CNN-LSTM only when the noise level reached $std = 30\%$.

On the other hand, compared to CLF-IT2FNN, the self-organizing learning mechanism achieved better model compactness, as evidenced by the reduced number of rules in both chaotic (Tables 1-3) and microgrid (Tables 4 and 5) prediction tasks. This is attributed to the proposed learning mechanism, which is able to dynamically adjust the rule generation and control rule redundancy. Specifically, the rule growing step (Algorithm 2) allows the model to continuously extract new rules for enhanced accuracy, while the rule removing step (Algorithm 3) enables the model to reduce redundant rules without compromising much accuracy. Through the iterative process of the learning mechanism (as described in Algorithm 1), an accurate model with a compact structure is achieved. Moreover, it is worth mentioning that the total number of rules generated by the MO scheme was less than (or equal to) that generated by PM, and more than that from SW. This is because PM involves optimizing multiple single-step prediction models, resulting in a higher total number of rules. In contrast, SW only needs to train one single-step prediction model, thus requiring the smallest number of rules.

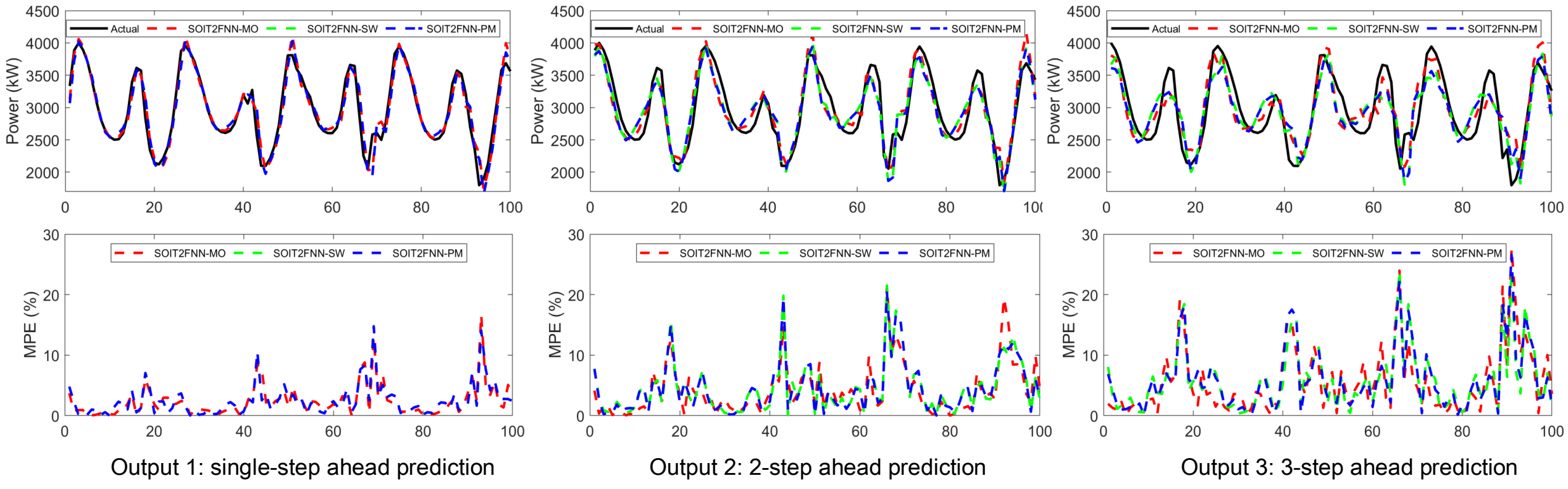

Figure 8: Performance comparisons on unmet power

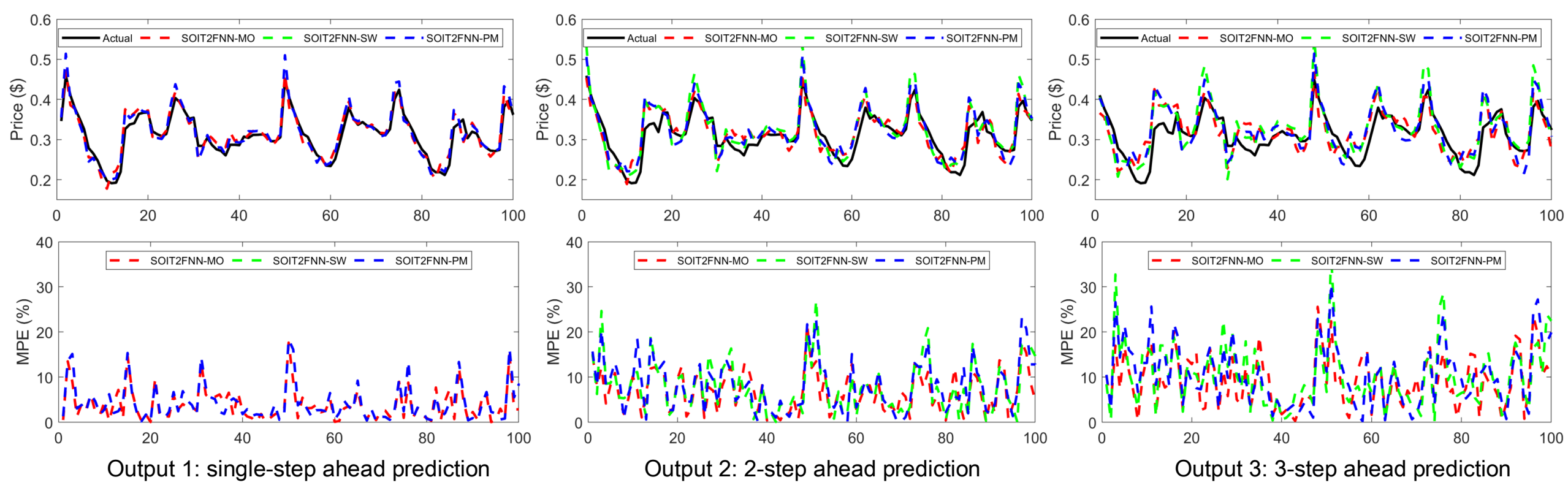

Figure 9: Performance comparisons on electricity price in Example 2

### *5.3. Sensitivity and Structure Evaluations*

This subsection first studies how the model structure and performance vary with the initial setting on the number of clusters in FCM. Table 6 lists the comparison results against different cluster numbers based on the clean datasets (see Table S6 in the supplementary materials for more details). Here, it can be observed that the impact of clustering initialization on model accuracy was minor. In fact, SOIT2FNN-MO tunes the antecedent parameters in stage 2 learning, helping alleviate the impact of clustering initialization. As for the model structure, increasing the number of clusters appeared to have negligible effect on chaotic and power prediction and have an irregular effect on price prediction. Overall, it would be a good choice to have fewer clusters in the pre-stage learning, as it can significantly reduce the computational costs without sacrificing the accuracy too much.

Table 4: Performance comparison on microgrid unmet power in Example 2

| Data | CNN-LSTM [11] | | | CLF-IT2NN [25] | | | **SOIT2FNN** | | |
|---|---|---|---|---|---|---|---|---|---|
| | SW | PM | MO | SW | PM | MO | SW | PM | **MO** |
| Training RMSE (Clean) | 0.07 | 0.06 | 0.05 | 0.11 | 0.10 | 0.10 | 0.09 | 0.09 | **0.08** |
| Test RMSE (Clean) | 0.07 | 0.07 | 0.07 | 0.11 | 0.09 | 0.09 | 0.09 | 0.09 | **0.08** |
| Test RMSE (std=10%) | 0.15 | 0.16 | 0.16 | 0.16 | 0.15 | 0.12 | 0.12 | 0.12 | **0.12** |
| Test RMSE (std=30%) | 0.33 | 0.33 | 0.32 | 0.28 | 0.27 | 0.26 | 0.25 | 0.25 | **0.25** |
| Training MPE (Clean) | 6.08 | 5.01 | 4.22 | 8.20 | 8.15 | 7.95 | 8.20 | 8.12 | **7.36** |
| Test MPE (Clean) | 6.37 | 5.83 | 6.01 | 8.36 | 8.08 | 8.15 | 8.21 | 8.10 | **7.26** |
| Test MPE (std=10%) | 14.7 | 15.1 | 15.4 | 14.8 | 13.3 | 12.1 | 11.8 | 11.8 | **11.4** |
| Test MPE (std=30%) | 34.9 | 34.3 | 34.6 | 31.4 | 27.9 | 28.3 | 26.5 | 26.6 | **26.3** |
| Rule | — | — | — | 1 | 1,1,2 | 3 | 1 | 1,1,1 | **3** |

Table 5: Performance comparison on the electricity price in Example 2

| Data | CNN-LSTM [11] | | | CLF-IT2NN [25] | | | **SOIT2FNN** | | |
|---|---|---|---|---|---|---|---|---|---|
| | SW | PM | MO | SW | PM | MO | SW | PM | **MO** |
| Training RMSE (Clean) | 0.06 | 0.05 | 0.05 | 0.09 | 0.08 | 0.08 | 0.08 | 0.8 | **0.07** |
| Test RMSE (Clean) | 0.06 | 0.06 | 0.06 | 0.08 | 0.08 | 0.08 | 0.07 | 0.07 | **0.07** |
| Test RMSE (std=10%) | 0.09 | 0.09 | 0.09 | 0.12 | 0.10 | 0.10 | 0.09 | 0.09 | **0.09** |
| Test RMSE (std=30%) | 0.18 | 0.19 | 0.21 | 0.19 | 0.17 | 0.17 | 0.16 | 0.16 | **0.16** |
| Training MPE (Clean) | 9.09 | 8.13 | 8.28 | 12.8 | 11.9 | 12.3 | 11.5 | 11.3 | **10.9** |
| Test MPE (Clean) | 10.2 | 9.26 | 9.75 | 12.0 | 11.7 | 11.7 | 11.7 | 11.5 | **11.2** |
| Test MPE (std=10%) | 15.8 | 15.9 | 16.2 | 20.0 | 18.4 | 17.6 | 17.1 | 16.5 | **16.4** |
| Test MPE (std=30%) | 40.6 | 44.8 | 48.1 | 42.5 | 37.9 | 38.3 | 37.9 | 37.2 | **36.7** |
| Rule | — | — | — | 1 | 1,1,2 | 3 | 1 | 1,1,1 | **2** |

Moreover, the structural design of the proposed SOIT2FNN-MO is also compared with the traditional structure of an IT2FNN (SIT2FNN: simplified interval type-2 neural fuzzy network) [52]. A transformation layer was added to the SIT2FNN once again to avoid the potential problem of vanished rule firing strength. The idea is to show if the newly introduced/modified layers in this paper can contribute to the model improvement. The results are given in Table 7; more details can be found in Tables S7 and S8 in the supplementary materials. It is evident that the new "Layer 4" and the modified "Layer 6" improved the prediction accuracy. The new "Layer 9" also demonstrated a positive impact on accuracy as it enhances temporal connections among multi-step predictions. Overall, the proposed SOIT2FNN-MO incorporating all these three layers achieved the best performance for multi-step time series predictions.

Furthermore, the effectiveness of the logistic operation (in Layer 5) is evaluated by removing it from the network. Unfortunately, the removal of the logistic operation resulted in training failure in simulations. Specifically, when the logistic operation is removed, $\overline{f}^{i,k} = \overline{R}^i R^k$ and $\underline{f}^{i,k} = \underline{R}^i R^k$ are obtained. In the microgrid case: 1) for a 9-input and 3-output model used for the power prediction example, $\overline{f}^{i,k}$ and $\underline{f}^{i,k}$ were both found below 1e-35; 2) for a 12-input and 3-output model, they were even less than 1e-100. When these extremely small values propagate into (12) and (13), they cause the "divide by zero" error. This demonstrates the rationality of the layer

Table 6: Performance comparison on clean datasets using different clustering numbers

| Data | No.=5 | No.=10 | No.=15 | No.=20 | No.=25 | No.=30 |
|---|---|---|---|---|---|---|
| Training RMSE (Clean):Chaotic | 0.027 | 0.028 | 0.025 | 0.024 | 0.026 | 0.023 |
| Test RMSE (Clean):Chaotic | 0.031 | 0.035 | 0.031 | 0.029 | 0.034 | 0.029 |
| Training MPE (Clean):Chaotic | 2.709 | 3.081 | 2.626 | 2.407 | 2.816 | 2.416 |
| Test MPE (Clean):Chaotic | 3.157 | 3.862 | 3.109 | 2.939 | 3.589 | 2.935 |
| Rule | 2 | 2 | 2 | 2 | 2 | 2 |
| Training RMSE (Clean):power | 0.084 | 0.085 | 0.083 | 0.083 | 0.082 | 0.081 |
| Test RMSE (Clean):power | 0.080 | 0.081 | 0.079 | 0.079 | 0.078 | 0.077 |
| Training MPE (Clean):power | 7.358 | 7.362 | 7.190 | 7.213 | 7.174 | 7.033 |
| Test MPE (Clean):power | 7.258 | 7.377 | 7.204 | 7.219 | 7.074 | 7.029 |
| Rule | 3 | 3 | 3 | 3 | 3 | 4 |
| Training RMSE (Clean):price | 0.077 | 0.074 | 0.076 | 0.070 | 0.074 | 0.081 |
| Test RMSE (Clean):price | 0.072 | 0.068 | 0.071 | 0.066 | 0.068 | 0.074 |
| Training MPE (Clean):price | 11.61 | 10.90 | 11.18 | 10.68 | 10.86 | 12.45 |
| Test MPE (Clean):price | 11.68 | 11.22 | 11.32 | 11.04 | 11.08 | 12.45 |
| Rule | 1 | 2 | 1 | 3 | 2 | 1 |

design and the importance of the logistic operation.

### 5.4. Interpretability Evaluation

Finally, the interpretability of the proposed SOTT2FNN-MO model is also evaluated using the chaotic time series data. As shown in Table 1, SOIT2FNN resulted in a 2-rule network. The firing strength of each rule is visualized across all 500 test samples, as shown in Fig. 10 (a) and (b). Here, $F^{i,k}$ (i.e., firing strength of the $i$-th rule with respect to the $k$-th output/prediction, as shown in (7)) is the output of Layer 5 and $F^i$ (if there is no Layer 4, i.e., traditional IT2FNN) is defined as:

$$F^i = [\underline{f}^i, \overline{f}^i], \quad i = 1, ..., M \tag{23}$$

where the lower bound ($\underline{f}^i$) and upper bound ($\overline{f}^i$) of $F^i$ can be expressed as:

$$\underline{f}^i = -\frac{1}{\log(\underline{R}^i)} = -\frac{1}{\sum_{j=1}^{n} log(\underline{\mu}_j^i(x_j))} \tag{24}$$

$$\overline{f}^i = -\frac{1}{\log(\overline{R}^i)} = -\frac{1}{\sum_{j=1}^{n} log(\overline{\mu}_j^i(x_j))} \tag{25}$$

It can be observed from the two figures (comparison between $F^i$ and $F^{i,k}$) that the new co-antecedent layer (Layer 4) gave a more concentrated distribution of firing strength in both rules. In most cases, rule 1 may have a greater impact than rule 2 as the firing strength of the former is generally larger. For more details, the 100th sample, $\tilde{x}$= [0.9810, 1.0408, 1.1740, 1.1630, 1.0770, 1.1085, 1.2025, 1.1471, 1.1177], was picked as an example for further analysis. The corresponding firing strengths ($F^i$ and $F^{i,k}$) of both rules are marked in Fig. 10 (a) and (b), respectively. It can be found that the firing strength of rule 1 is much higher than that of rule 2. Moreover, the firing strengths for all three predictions are very close in both rules. This is because SOIT2FNN-MO is designed for multi-step time-series predictions, and there are clear temporal dependencies (very close values) among the three outputs. Unlike Layer 2 and Layer 3 (which work for all multi-step predictions), Layer 4 is an output-oriented layer (i.e., it only plays an important role in one of the model outputs). In this manner, the presence of Layer 4 allows for a better interpretation of each prediction in multi-step time-series forecasting. It should be noted that while Layer 4 may also affect the predictions at other time steps due to the presence of the final link layer in the network; such effect is typically minimal. This is because link values $l$ is generally very small, as elaborated later.

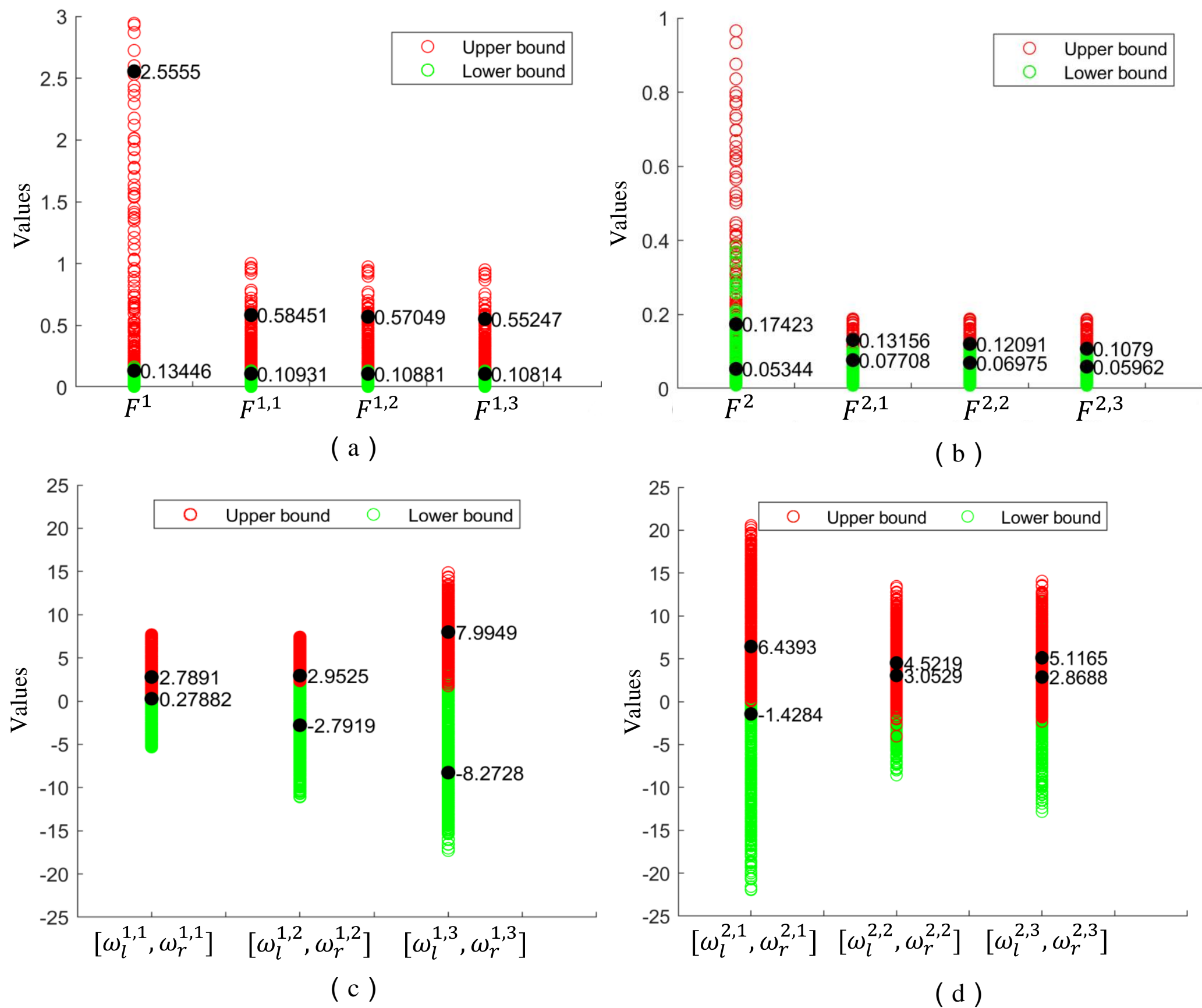

Figure 10: The firing strengths of rule 1 (a) and rule 2 (b); the outputs of the modified consequent layer for rule 1 (c) and rule 2 (d)

The modified consequent (Layer 6) is then analyzed by visualizing its outputs for each rule on all 500 test samples, as shown in Fig. 10 (c) and (d). Similar to the role of Layer 4, the modified Layer 6 can present the behaviours of the prediction at each time step. However, unlike the patterns showed in the rule firing strength, the distributions of the three outputs vary significantly between each rule. This can be explained by the underpinning principle of TSK-type fuzzy models. In detail, the rule antecedent (IF part) is used to partition the entire input space into multiple local fuzzy regions, enabling localized representation of input patterns [67]. Then, the rule consequent (THEN part) characterizes the system's behaviours within each local region, capturing the underlying dynamics of the model [68]. For multi-step prediction problems where the inputs are also time series, the local regions defined by rule antecedents can be quite similar - see Fig. 10 (a) and (b). However, the behaviour of each rule may present different patterns based

Table 7: Performance evaluation for each modified/added layer (models trained with clean data )

| Data | SIT2FNN [52] | Proposed | S+L4 | S+L6 | S+L9 |
|---|---|---|---|---|---|
| Test RMSE (Clean):Chaotic | 0.041 | 0.031 | 0.043 | 0.031 | 0.039 |
| Test RMSE (std=10%):Chaotic | 0.173 | 0.165 | 0.169 | 0.176 | 0.173 |
| Test RMSE (std=30%):Chaotic | 0.529 | 0.496 | 0.492 | 0.503 | 0.490 |
| Test MPE (Clean):Chaotic | 4.548 | 3.157 | 4.216 | 3.315 | 4.292 |
| Test MPE (std=10%):Chaotic | 20.27 | 18.61 | 18.66 | 19.79 | 20.00 |
| Test MPE (std=30%):Chaotic | 73.28 | 67.20 | 69.38 | 69.32 | 66.63 |
| Rule | 3 | 2 | 3 | 2 | 3 |
| Test RMSE (Clean):power | 0.086 | 0.080 | 0.083 | 0.089 | 0.086 |
| Test RMSE (std=10%):power | 0.120 | 0.116 | 0.118 | 0.117 | 0.117 |
| Test RMSE (std=30%):power | 0.265 | 0.249 | 0.253 | 0.250 | 0.250 |
| Test MPE (Clean):power | 8.152 | 7.258 | 7.758 | 8.050 | 8.154 |
| Test MPE (std=10%):power | 12.65 | 11.38 | 12.54 | 11.58 | 11.69 |
| Test MPE (std=30%):power | 28.26 | 26.27 | 26.19 | 25.65 | 25.87 |
| Rule | 2 | 3 | 3 | 1 | 2 |
| Test RMSE (Clean):price | 0.083 | 0.068 | 0.073 | 0.074 | 0.073 |
| Test RMSE (std=10%):price | 0.092 | 0.086 | 0.088 | 0.088 | 0.088 |
| Test RMSE (std=30%):price | 0.169 | 0.160 | 0.164 | 0.161 | 0.160 |
| Test MPE (Clean):price | 12.79 | 11.22 | 11.79 | 11.49 | 11.36 |
| Test MPE (std=10%):price | 17.47 | 16.43 | 16.43 | 17.73 | 16.08 |
| Test MPE (std=30%):price | 36.83 | 36.67 | 36.74 | 38.43 | 36.96 |
| Rule | 2 | 2 | 2 | 2 | 2 |

**Note**: S+L4, S+L6 and S+L9 represent the model of SIT2FNN [52] with Layer 4, Layer 6 and Layer 9, respectively

on the rule consequent - see Fig. 10 (c) and (d).

In addition to the parameters in the rule antecedents and consequents mentioned above, other parameters also showed good interpretability. Specifically, the parameter $q_o^k$ in Layer 8 indicates the importance of the interval boundaries $y_l^k$ and $y_r^k$ for the *k-th* prediction, while the link parameter $l$ in Layer 9 demonstrates the significance between the prediction at the previous and current time steps for the final result. For chaotic time series prediction, the trained parameters obtained were: $q_o^1 = 0.578, q_o^2 = 0.726, q_o^3 = 0.429, l = 0.124$. This suggests that the predictions at the first two steps are more likely to be influenced by the lower interval boundary. In contrast, the upper interval boundary influences the last prediction more. The predicted value from Layer 8 contributes to 87.6% of the final prediction, while the rest is attributed to predictions from previous time steps.

### *5.5. Discussion*

This paper primarily focuses on improving the IT2FNN architecture for multi-step time series forecasting problems, as well as the learning mechanism for automating the model construction process. All nonlinear and linear network parameters including both the rule antecedent and consequent parameters, together with the model structure, are optimized by the proposed two-stage learning mechanism. Moreover, as demonstrated in the results, the proposed SOIT2FNN-MO model is shown to generate more accurate predictions compared to alternative models (especially when the data is noisy).

The proposed model offers both theoretical and practical advancements in multi-step time series prediction. Theoretically, it enhances the interpretability of IT2FNN by introducing a novel co-antecedent layer and a modified consequent layer, where each model output is governed by a different rule representation. Additionally, the new link layer effectively captures temporal dependencies between predictions at multiple time steps, addressing a crucial limitation of IT2FNN-based forecasting models. The added transformation layer further empowers the model to handle high-dimensional inputs by mitigating the issue of vanishing rule strength. These architectural innovations, combined with the two-stage learning mechanism, contribute to the theoretical development of IT2FNN and its applications to time series forecasting problems.

From a practical perspective, SOIT2FNN-MO significantly improves forecasting accuracy under both clean and noisy conditions, as demonstrated through extensive experiments on chaotic and microgrid datasets. The ability to automatically generate fuzzy rules from data avoids manual structure design, making it scalable to real-world applications, such as energy load forecasting and environmental monitoring [69]. Furthermore, the enhanced interpretability provides users with more insights and trust into the prediction outcomes. Despite these advantages, training the proposed model requires a significant computational demand due to the nine-layer structure and self-organizing mechanism. Further research is needed to enhance the computational efficiency, particularly for practical applications involving large datasets.

## 6. CONCLUSION

Model accuracy and interpretability in multi-step time series prediction under data uncertainty remains a fundamental challenge. To effectively tackle this challenge, this paper proposed a self-organizing interval type-2 fuzzy neural network with multiple outputs (SOIT2FNN-MO). On the one hand, a nine-layer deep neural architecture was designed for interval type-2 fuzzy systems to address jointly the problems of model accuracy, interpretability and uncertainty handling. On the other hand, a two-stage, self-organizing learning mechanism was developed to automatically generate fuzzy rules (i.e., the model structure) and optimize both antecedent and consequent rule parameters. Simulations on chaotic time series and microgrid variables, such as electricity price and unmet power, demonstrate that the proposed SOIT2FNN-MO consistently outperforms existing approaches, delivering enhanced accuracy and interpretability. The observed accuracy improvements, ranging from 1.6% to 30%, are affected by the degree of noise added to the data (e.g., standard deviations of 10% and 30%). Despite this achievement, the proposed approach involves a high computational demand. Future research will focus on reducing model complexity and improving computational efficiency, such as using model compression techniques or developing hybrid learning approaches by incorporating domain knowledge.

# Supplementary Materials - A Self-organizing Interval Type-2 Fuzzy Neural Network for Multi-Step Time Series Prediction

Fulong Yao[a,*], Wanqing Zhao[a], Matthew Forshaw[a], Yang Song[b]

[a]*School of Computing, Newcastle University, Newcastle Upon Tyne, NE4 5TG, UK*

[b]*School of Mechanical and Electrical Engineering and Automation, Shanghai University, Shanghai, 200444, China*

---

**Abstract**

This is the supplementary document of the paper entitled " A Self-organizing Interval Type-2 Fuzzy Neural Network for Multi-Step Time Series Prediction," which has been accepted for publication in *Applied Soft Computing*. Section 1 gives the derivations of the gradient descent optimization in the learning of the proposed SOIT2FNN-MO, while Section 2 provides additional simulations in chaotic and two microgrid time series prediction problems.

---

## 1. SUPPLEMENT ON DERIVATIONS

### 1.1. Derivation of Gradients for (21)

The gradient derivations of the network parameters (i.e., $c_j^{i,k}$, $s_j^{i,k}$, $m_{1,j}^i$, $m_{2,j}^i$, $\sigma_j^i$, $m_j^k$, $\sigma_j^k$, $q_l^k$, $q_r^k$, $q_o^k$, $l$) can be expressed as follows:

$$\frac{\partial E}{\partial c_j^{i,k}} = \frac{\partial E}{\partial y^{k'}}\left(\frac{\partial y^{k'}}{\partial y_l^k}\frac{\partial y_l^k}{\partial w_l^{i,k}}\frac{\partial w_l^{i,k}}{\partial c_j^{i,k}} + \frac{\partial y^{k'}}{\partial y_r^k}\frac{\partial y_r^k}{\partial w_r^{i,k}}\frac{\partial w_r^{i,k}}{\partial c_j^{i,k}}\right) \tag{S1}$$

$$\frac{\partial E}{\partial s_j^{i,k}} = \frac{\partial E}{\partial y^{k'}}\left(\frac{\partial y^{k'}}{\partial y_l^k}\frac{\partial y_l^k}{\partial w_l^{i,k}}\frac{\partial w_l^{i,k}}{\partial s_j^{i,k}} + \frac{\partial y^{k'}}{\partial y_r^k}\frac{\partial y_r^k}{\partial w_r^{i,k}}\frac{\partial w_r^{i,k}}{\partial s_j^{i,k}}\right) \tag{S2}$$

$$\frac{\partial E}{\partial m_{1,j}^i} = \sum_{k=1}^{K}\left[\frac{\partial E}{\partial y^{k'}}\left(\frac{\partial y^{k'}}{\partial y_l^k}\frac{\partial y_l^k}{\partial \overline{f}^{i,k}} + \frac{\partial y^{k'}}{\partial y_r^k}\frac{\partial y_r^k}{\partial \overline{f}^{i,k}}\right)\frac{\partial \overline{f}^{i,k}}{\partial m_{1,j}^i}\right] + \sum_{k=1}^{K}\left[\frac{\partial E}{\partial y^{k'}}\left(\frac{\partial y^{k'}}{\partial y_l^k}\frac{\partial y_l^k}{\partial \underline{f}^{i,k}} + \frac{\partial y^{k'}}{\partial y_r^k}\frac{\partial y_r^k}{\partial \underline{f}^{i,k}}\right)\frac{\partial \underline{f}^{i,k}}{\partial m_{1,j}^i}\right] \tag{S3}$$

$$\frac{\partial E}{\partial m_{2,j}^i} = \sum_{k=1}^{K}\left[\frac{\partial E}{\partial y^{k'}}\left(\frac{\partial y^{k'}}{\partial y_l^k}\frac{\partial y_l^k}{\partial \overline{f}^{i,k}} + \frac{\partial y^{k'}}{\partial y_r^k}\frac{\partial y_r^k}{\partial \overline{f}^{i,k}}\right)\frac{\partial \overline{f}^{i,k}}{\partial m_{2,j}^i}\right] + \sum_{k=1}^{K}\left[\frac{\partial E}{\partial y^{k'}}\left(\frac{\partial y^{k'}}{\partial y_l^k}\frac{\partial y_l^k}{\partial \underline{f}^{i,k}} + \frac{\partial y^{k'}}{\partial y_r^k}\frac{\partial y_r^k}{\partial \underline{f}^{i,k}}\right)\frac{\partial \underline{f}^{i,k}}{\partial m_{2,j}^i}\right] \tag{S4}$$

$$\frac{\partial E}{\partial \sigma_j^i} = \sum_{k=1}^{K}\left[\frac{\partial E}{\partial y^{k'}}\left(\frac{\partial y^{k'}}{\partial y_l^k}\frac{\partial y_l^k}{\partial \overline{f}^{i,k}} + \frac{\partial y^{k'}}{\partial y_r^k}\frac{\partial y_r^k}{\partial \overline{f}^{i,k}}\right)\frac{\partial \overline{f}^{i,k}}{\partial \sigma_j^i}\right] + \sum_{k=1}^{K}\left[\frac{\partial E}{\partial y^{k'}}\left(\frac{\partial y^{k'}}{\partial y_l^k}\frac{\partial y_l^k}{\partial \underline{f}^{i,k}} + \frac{\partial y^{k'}}{\partial y_r^k}\frac{\partial y_r^k}{\partial \underline{f}^{i,k}}\right)\frac{\partial \underline{f}^{i,k}}{\partial \sigma_j^i}\right] \tag{S5}$$

---

*Corresponding author: Fulong Yao, f.yao3@newcastle.ac.uk

$$\frac{\partial E}{\partial m_j^k} = \sum_{i=1}^{M}[\frac{\partial E}{\partial y^{k\prime}}(\frac{\partial y^{k\prime}}{\partial y_l^k}\frac{\partial y_l^k}{\partial \overline{f}^{i,k}} + \frac{\partial y^{k\prime}}{\partial y_r^k}\frac{\partial y_r^k}{\partial \overline{f}^{i,k}})\frac{\partial \overline{f}^{i,k}}{\partial m_j^k}] + \sum_{i=i}^{M}[\frac{\partial E}{\partial y^{k\prime}}(\frac{\partial y^{k\prime}}{\partial y_l^k}\frac{\partial y_l^k}{\partial \underline{f}^{i,k}} + \frac{\partial y^{k\prime}}{\partial y_r^k}\frac{\partial y_r^k}{\partial \underline{f}^{i,k}})\frac{\partial \underline{f}^{i,k}}{\partial m_j^k}] \tag{S6}$$

$$\frac{\partial E}{\partial \sigma_j^k} = \sum_{i=1}^{M}[\frac{\partial E}{\partial y^{k\prime}}(\frac{\partial y^{k\prime}}{\partial y_l^k}\frac{\partial y_l^k}{\partial \overline{f}^{i,k}} + \frac{\partial y^{k\prime}}{\partial y_r^k}\frac{\partial y_r^k}{\partial \overline{f}^{i,k}})\frac{\partial \overline{f}^{i,k}}{\partial \sigma_j^k}] + \sum_{i=i}^{M}[\frac{\partial E}{\partial y^{k\prime}}(\frac{\partial y^{k\prime}}{\partial y_l^k}\frac{\partial y_l^k}{\partial \underline{f}^{i,k}} + \frac{\partial y^{k\prime}}{\partial y_r^k}\frac{\partial y_r^k}{\partial \underline{f}^{i,k}})\frac{\partial \underline{f}^{i,k}}{\partial \sigma_j^k}] \tag{S7}$$

$$\frac{\partial E}{\partial q_l^k} = \frac{\partial E}{\partial y^{k\prime}}\frac{\partial y^{k\prime}}{\partial y_l^k}\frac{\partial y_l^k}{\partial q_l^k} \tag{S8}$$

$$\frac{\partial E}{\partial q_r^k} = \frac{\partial E}{\partial y^{k\prime}}\frac{\partial y^{k\prime}}{\partial y_r^k}\frac{\partial y_r^k}{\partial q_r^k} \tag{S9}$$

$$\frac{\partial E}{\partial q_o^k} = \frac{\partial E}{\partial y^{k\prime}}\frac{\partial y^{k\prime}}{\partial q_o^k} \tag{S10}$$

$$\frac{\partial E}{\partial l} = \sum_{\mathbf{k}=1}^{K}\frac{\partial E}{\partial y^{\mathbf{k}}}\frac{\partial y^{\mathbf{k}}}{\partial l} \tag{S11}$$

where

$$\frac{\partial E}{\partial y^{k\prime}} = \sum_{\mathbf{k}=k}^{K}(\frac{\partial E}{\partial y^{\mathbf{k}}}\frac{\partial y^{\mathbf{k}}}{\partial y^{k\prime}}) \tag{S12}$$

$$\frac{\partial E}{\partial y^{\mathbf{k}}} = y^k - y_a^k \tag{S13}$$

$$\frac{\partial y^{\mathbf{k}}}{\partial y^{k\prime}} = l^{\mathbf{k}-k}(1-l) \qquad \text{(see - Supplement 1.2 below)} \tag{S14}$$

and

$$\frac{\partial y^{k\prime}}{\partial y_l^k} = q_o^k \tag{S15}$$

$$\frac{\partial y^{k\prime}}{\partial y_r^k} = (1-q_o^k) \tag{S16}$$

$$\frac{\partial y_l^k}{\partial w_l^{i,k}}\frac{\partial w_l^{i,k}}{\partial c_j^{i,k}} = \frac{(1-q_l^k)\underline{f}^{i,k} + q_l^k\overline{f}^{i,k}}{\sum_{i=1}^{M}(\underline{f}^{i,k} + \overline{f}^{i,k})}x_j \tag{S17}$$

$$\frac{\partial y_r^k}{\partial w_r^{i,k}}\frac{\partial w_r^{i,k}}{\partial c_j^{i,k}} = \frac{(1-q_r^k)\underline{f}^{i,k} + q_r^k\overline{f}^{i,k}}{\sum_{i=1}^{M}(\underline{f}^{i,k} + \overline{f}^{i,k})}x_j \tag{S18}$$

$$\frac{\partial y_l^k}{\partial w_l^{i,k}}\frac{\partial w_l^{i,k}}{\partial s_j^{i,k}} = -\frac{(1-q_l^k)\underline{f}^{i,k} + q_l^k\overline{f}^{i,k}}{\sum_{i=1}^{M}(\underline{f}^{i,k} + \overline{f}^{i,k})}|x_j| \tag{S19}$$

$$\frac{\partial y_r^k}{\partial w_r^{i,k}}\frac{\partial w_r^{i,k}}{\partial s_j^{i,k}} = \frac{(1-q_r^k)\underline{f}^{i,k} + q_r^k\overline{f}^{i,k}}{\sum_{i=1}^{M}(\underline{f}^{i,k} + \overline{f}^{i,k})}|x_j| \tag{S20}$$

$$\frac{\partial y^{k\prime}}{\partial y_l^k}\frac{\partial y_l^k}{\partial q_l^k} = \frac{\sum_{i=1}^{M}(\overline{f}^{i,k} - \underline{f}^{i,k})w_l^{i,k}}{\sum_{i=1}^{M}(\underline{f}^{i,k} + \overline{f}^{i,k})} \tag{S21}$$

$$\frac{\partial y^{k\prime}}{\partial y_r^k}\frac{\partial y_r^k}{\partial q_r^k} = \frac{\sum_{i=1}^{M}(\overline{f}^{i,k} - \underline{f}^{i,k})w_r^{i,k}}{\sum_{i=1}^{M}(\underline{f}^{i,k} + \overline{f}^{i,k})} \tag{S22}$$

$$\frac{\partial y^{k\prime}}{\partial q_o^k} = y_l^k - y_r^k \tag{S23}$$

$$\frac{\partial y_l^k}{\partial \overline{f}^{i,k}} = \frac{q_l^k w_l^{i,k} - y_l^k}{\sum_{i=1}^{M}(\underline{f}^{i,k} + \overline{f}^{i,k})} \tag{S24}$$

$$\frac{\partial y_r^k}{\partial \overline{f}^{i,k}} = \frac{q_r^k w_r^{i,k} - y_r^k}{\sum_{i=1}^{M}(\underline{f}^{i,k} + \overline{f}^{i,k})} \tag{S25}$$

$$\frac{\partial y_l^k}{\partial \underline{f}^{i,k}} = \frac{(1 - q_l^k)w_l^{i,k} - y_l^k}{\sum_{i=1}^{M}(\underline{f}^{i,k} + \overline{f}^{i,k})} \tag{S26}$$

$$\frac{\partial y_r^k}{\partial \underline{f}^{i,k}} = \frac{(1 - q_r^k)w_r^{i,k} - y_r^k}{\sum_{i=1}^{M}(\underline{f}^{i,k} + \overline{f}^{i,k})} \tag{S27}$$

$$\frac{\partial y^{\mathbf{k}}}{\partial l} = \sum_{k=1}^{\mathbf{k}}(\mathbf{k} - k + 1)l^{\mathbf{k}-k}(y^{k-1\prime} - y^k) \quad \text{(see - later Supplement 1.2 below)} \tag{S28}$$

$$\frac{\partial \overline{f}^{i,k}}{\partial m_{1,j}^i} = \frac{\partial \overline{f}^{i,k}}{\partial \overline{R}^i}\frac{\partial \overline{R}^i}{\partial m_{1,j}^i} = \begin{cases} \left(\overline{f}^{i,k}\right)^2 \times \dfrac{x_j - m_{1,j}^i}{(\sigma_j^i)^2}, & if\ x_j \leq m_{1,j}^i; \\ 0, & otherwise. \end{cases} \tag{S29}$$

$$\frac{\partial \underline{f}^{i,k}}{\partial m_{1,j}^i} = \frac{\partial \underline{f}^{i,k}}{\partial \underline{R}^i}\frac{\partial \underline{R}^i}{\partial m_{1,j}^i} = \begin{cases} \left(\underline{f}^{i,k}\right)^2 \times \dfrac{x_j - m_{1,j}^i}{(\sigma_j^i)^2}, & if\ x_j > \dfrac{m_{1,j}^i + m_{2,j}^i}{2}; \\ 0, & otherwise. \end{cases} \tag{S30}$$

$$\frac{\partial \overline{f}^{i,k}}{\partial m_{2,j}^i} = \frac{\partial \overline{f}^{i,k}}{\partial \overline{R}^i}\frac{\partial \overline{R}^i}{\partial m_{2,j}^i} = \begin{cases} \left(\overline{f}^{i,k}\right)^2 \times \dfrac{x_j - m_{2,j}^i}{(\sigma_j^i)^2}, & if\ x_j > m_{2,j}^i; \\ 0, & otherwise. \end{cases} \tag{S31}$$

$$\frac{\partial \underline{f}^{i,k}}{\partial m_{2,j}^i} = \frac{\partial \underline{f}^{i,k}}{\partial \underline{R}^i}\frac{\partial \underline{R}^i}{\partial m_{2,j}^i} = \begin{cases} \left(\underline{f}^{i,k}\right)^2 \times \dfrac{x_j - m_{2,j}^i}{(\sigma_j^i)^2}, & if\ x_j \leq \dfrac{m_{1,j}^i + m_{2,j}^i}{2}; \\ 0, & otherwise. \end{cases} \tag{S32}$$

$$\frac{\partial \overline{f}^{i,k}}{\partial \sigma_j^i} = \frac{\partial \overline{f}^{i,k}}{\partial \overline{R}^i}\frac{\partial \overline{R}^i}{\partial \sigma_j^i} = \begin{cases} \left(\overline{f}^{i,k}\right)^2 \times \dfrac{(x_j - m_{1,j}^i)^2}{(\sigma_j^i)^3}, & if\ x_j < m_{1,j}^i; \\ \left(\overline{f}^{i,k}\right)^2 \times \dfrac{(x_j - m_{2,j}^i)^2}{(\sigma_j^i)^3}, & if\ x_j > m_{2,j}^i; \\ 0, & otherwise. \end{cases} \tag{S33}$$

$$\frac{\partial \underline{f}^{i,k}}{\partial \sigma_j^i} = \frac{\partial \underline{f}^{i,k}}{\partial \underline{R}^i}\frac{\partial \underline{R}^i}{\partial \sigma_j^i} = \begin{cases} \left(\underline{f}^{i,k}\right)^2 \times \dfrac{(x_j - m_{2,j}^i)^2}{(\sigma_j^i)^3}, & if\ x_j \le \dfrac{m_{1,j}^i + m_{2,j}^i}{2}; \\ \left(\underline{f}^{i,k}\right)^2 \times \dfrac{(x_j - m_{1,j}^i)^2}{(\sigma_j^i)^3}, & if\ x_j > \dfrac{m_{1,j}^i + m_{2,j}^i}{2}. \end{cases} \tag{S34}$$

$$\frac{\partial \overline{f}^{i,k}}{\partial m_j^k} = \frac{\partial \overline{f}^{i,k}}{\partial R^k}\frac{\partial R^k}{\partial m_j^k} = \left(\overline{f}^{i,k}\right)^2 \times \frac{x_j - m_j^k}{(\sigma_j^k)^2} \tag{S35}$$

$$\frac{\partial \underline{f}^{i,k}}{\partial m_j^k} = \frac{\partial \underline{f}^{i,k}}{\partial R^k}\frac{\partial R^k}{\partial m_j^k} = \left(\underline{f}^{i,k}\right)^2 \times \frac{x_j - m_j^k}{(\sigma_j^k)^2} \tag{S36}$$

$$\frac{\partial \overline{f}^{i,k}}{\partial \sigma_j^k} = \frac{\partial \overline{f}^{i,k}}{\partial R^k}\frac{\partial R^k}{\partial \sigma_j^k} = \left(\overline{f}^{i,k}\right)^2 \times \frac{(x_j - m_j^k)^2}{(\sigma_j^k)^3} \tag{S37}$$

$$\frac{\partial \underline{f}^{i,k}}{\partial \sigma_j^k} = \frac{\partial \underline{f}^{i,k}}{\partial R^k}\frac{\partial R^k}{\partial \sigma_j^k} = \left(\underline{f}^{i,k}\right)^2 \times \frac{(x_j - m_j^k)^2}{(\sigma_j^k)^3} \tag{S38}$$

The derivations of $\frac{\partial \overline{f}^{i,k}}{\partial m_{1,j}^i}$, $\frac{\partial \underline{f}^{i,k}}{\partial m_{1,j}^i}$, $\frac{\partial \overline{f}^{i,k}}{\partial m_{2,j}^i}$, $\frac{\partial \underline{f}^{i,k}}{\partial m_{2,j}^i}$, $\frac{\partial \overline{f}^{i,k}}{\partial \sigma_j^i}$, $\frac{\partial \underline{f}^{i,k}}{\partial \sigma_j^i}$, $\frac{\partial \overline{f}^{i,k}}{\partial m_j^k}$, $\frac{\partial \underline{f}^{i,k}}{\partial m_j^k}$, $\frac{\partial \overline{f}^{i,k}}{\partial \sigma_j^k}$ and $\frac{\partial \underline{f}^{i,k}}{\partial \sigma_j^k}$ can be found in Supplement 1.3 below.

### *1.2. Derivation of* $\frac{\partial y^{\mathbf{k}}}{\partial y^{k'}}$

Based on (15), the following can be derived:

$$\begin{aligned} & y^1 = (1-l)y^{1'} + l x_n \\ & y^2 = (1-l)y^{2'} + l y^1 = (1-l)y^{2'} + l(1-l)y^{1'} + l^2 x_n \\ & y^3 = (1-l)y^{3'} + l y^2 = (1-l)y^{3'} + l(1-l)y^{2'} + l^2(1-l)y^{1'} + l^3 x_n \\ & .... \\ & y^k = (1-l)y^{k'} + l y^k = (1-l)y^{k'} + ... + l^{k-1}(1-l)y^{1'} + l^k x_n \end{aligned} \tag{S39}$$

Then, it can be expressed explicitly as:

$$y^{\mathbf{k}} = \sum_{k=1}^{\mathbf{k}} l^{\mathbf{k}-k}(1-l)y^{k'} + l^{\mathbf{k}} x_n \tag{S40}$$

where $k, \mathbf{k} \in N^+$ and $k \le \mathbf{k}$.

Therefore, the following holds:

$$\frac{\partial y^{\mathbf{k}}}{\partial y^{k'}} = l^{\mathbf{k}-k}(1-l) \tag{S41}$$

Further considering,

$$
\begin{aligned}
\frac{\partial y^1}{\partial l} &= -y^{1'} + x_n \\
\frac{\partial y^2}{\partial l} &= -y^{2'} + y^{1'} - 2ly^{1'} + 2lx_n \\
\frac{\partial y^3}{\partial l} &= -y^{3'} + y^{3'} - 2ly^{3'} + 2ly^{1'} - 3l^2y^{1'} + 3l^2x_n \\
&\;.... \\
\frac{\partial y^k}{\partial l} &= -y^{k'} + y^{k-1'} - ... - kl^{k-1}y^{1'} + kl^{k-1}x_n
\end{aligned}
\tag{S42}
$$

and let $x_n = y^{0'}$, then the following can be obtained:

$$
\frac{\partial y^{\mathbf{k}}}{\partial l} = \sum_{k=1}^{\mathbf{k}} (\mathbf{k} - k + 1) l^{\mathbf{k}-k} (y^{k-1'} - y^k)
\tag{S43}
$$

### *1.3. Derivation of* $\frac{\partial \overline{f}^{i,k}}{\partial m^i_{1,j}}$, $\frac{\partial \underline{f}^{i,k}}{\partial m^i_{1,j}}$, $\frac{\partial \overline{f}^{i,k}}{\partial m^i_{2,j}}$, $\frac{\partial \underline{f}^{i,k}}{\partial m^i_{2,j}}$, $\frac{\partial \overline{f}^{i,k}}{\partial \sigma^i_j}$, $\frac{\partial \underline{f}^{i,k}}{\partial \sigma^i_j}$, $\frac{\partial \overline{f}^{i,k}}{\partial m^k_j}$, $\frac{\partial \underline{f}^{i,k}}{\partial m^k_j}$, $\frac{\partial \overline{f}^{i,k}}{\partial \sigma^k_j}$, $\frac{\partial \underline{f}^{i,k}}{\partial \sigma^k_j}$

According to (8) and (9), we have:

$$
\frac{\partial \overline{f}^{i,k}}{\partial \overline{R}^i} = \frac{1}{\overline{R}^i R^k (log(\overline{R}^i R^k))^2} R^k = \frac{\left(\overline{f}^{i,k}\right)^2}{\overline{R}^i}
\tag{S44}
$$

$$
\frac{\partial \underline{f}^{i,k}}{\partial \underline{R}^i} = \frac{1}{\underline{R}^i R^k (log(\underline{R}^i R^k))^2} R^k = \frac{\left(\underline{f}^{i,k}\right)^2}{\underline{R}^i}
\tag{S45}
$$

$$
\frac{\partial \overline{f}^{i,k}}{\partial R^k} = \frac{1}{\overline{R}^i R^k (log(\overline{R}^i R^k))^2} \overline{R}^i = \frac{\left(\overline{f}^{i,k}\right)^2}{R^k}
\tag{S46}
$$

$$
\frac{\partial \underline{f}^{i,k}}{\partial R^k} = \frac{1}{\underline{R}^i R^k (log(\underline{R}^i R^k))^2} \underline{R}^i = \frac{\left(\underline{f}^{i,k}\right)^2}{R^k}
\tag{S47}
$$

According to (2) - (6), these can be further derived:

$$
\frac{\partial \overline{R}^i}{\partial m^i_{1,j}} = \frac{\partial \overline{R}^i}{\partial \overline{\mu}^i_j} \frac{\partial \overline{\mu}^i_j}{\partial m^i_{1,j}} = \begin{cases} \overline{R}^i \times \dfrac{x_j - m^i_{1,j}}{(\sigma^i_j)^2}, & if\ x_j \leq m^i_{1,j}; \\ 0, & otherwise. \end{cases}
\tag{S48}
$$

$$
\frac{\partial \underline{R}^i}{\partial m^i_{1,j}} = \frac{\partial \underline{R}^i}{\partial \underline{\mu}^i_j} \frac{\partial \underline{\mu}^i_j}{\partial m^i_{1,j}} = \begin{cases} \underline{R}^i \times \dfrac{x_j - m^i_{1,j}}{(\sigma^i_j)^2}, & if\ x_j > \dfrac{m^i_{1,j} + m^i_{2,j}}{2}; \\ 0, & otherwise. \end{cases}
\tag{S49}
$$

$$
\frac{\partial \overline{R}^i}{\partial m^i_{2,j}} = \frac{\partial \overline{R}^i}{\partial \overline{\mu}^i_j} \frac{\partial \overline{\mu}^i_j}{\partial m^i_{2,j}} = \begin{cases} \overline{R}^i \times \dfrac{x_j - m^i_{2,j}}{(\sigma^i_j)^2}, & if\ x_j > m^i_{2,j}; \\ 0, & otherwise. \end{cases}
\tag{S50}
$$

$$\frac{\partial \underline{R}^i}{\partial m_{2,j}^i} = \frac{\partial \underline{R}^i}{\partial \underline{\mu}_j^i}\frac{\partial \underline{\mu}_j^i}{\partial m_{2,j}^i} = \begin{cases} \underline{R}^i \times \dfrac{x_j - m_{2,j}^i}{(\sigma_j^i)^2}, & if\ x_j \leq \dfrac{m_{1,j}^i + m_{2,j}^i}{2}; \\ 0, & otherwise. \end{cases} \tag{S51}$$

$$\frac{\partial \overline{R}^i}{\partial \sigma_j^i} = \frac{\partial \overline{R}^i}{\partial \overline{\mu}_j^i}\frac{\partial \overline{\mu}_j^i}{\partial \sigma_j^i} = \begin{cases} \overline{R}^i \times \dfrac{(x_j - m_{1,j}^i)^2}{(\sigma_j^i)^3}, & if\ x_j < m_{1,j}^i; \\ \overline{R}^i \times \dfrac{(x_j - m_{2,j}^i)^2}{(\sigma_j^i)^3}, & if\ x_j > m_{2,j}^i; \\ 0, & otherwise. \end{cases} \tag{S52}$$

$$\frac{\partial \underline{R}^i}{\partial \sigma_j^i} = \frac{\partial \underline{R}^i}{\partial \underline{\mu}_j^i}\frac{\partial \underline{\mu}_j^i}{\partial \sigma_j^i} = \begin{cases} \underline{R}^i \times \dfrac{(x_j - m_{2,j}^i)^2}{(\sigma_j^i)^3}, & if\ x_j \leq \dfrac{m_{1,j}^i + m_{2,j}^i}{2}; \\ \underline{R}^i \times \dfrac{(x_j - m_{1,j}^i)^2}{(\sigma_j^i)^3}, & if\ x_j > \dfrac{m_{1,j}^i + m_{2,j}^i}{2}. \end{cases} \tag{S53}$$

$$\frac{\partial R^k}{\partial m_j^k} = \frac{\partial R^k}{\partial \mu_j^k}\frac{\partial \mu_j^k}{\partial m_j^k} = R^k \times \frac{x_j - m_j^k}{(\sigma_j^k)^2} \tag{S54}$$

$$\frac{\partial R^k}{\partial \sigma_j^k} = \frac{\partial R^k}{\partial \mu_j^k}\frac{\partial \mu_j^k}{\partial \sigma_j^k} = R^k \times \frac{(x_j - m_j^k)^2}{(\sigma_j^k)^3} \tag{S55}$$

Based on the above, finally the following quantities can be obtained:

$$\frac{\partial \overline{f}^{i,k}}{\partial m_{1,j}^i} = \frac{\partial \overline{f}^{i,k}}{\partial \overline{R}^i}\frac{\partial \overline{R}^i}{\partial m_{1,j}^i} = \begin{cases} \left(\overline{f}^{i,k}\right)^2 \times \dfrac{x_j - m_{1,j}^i}{(\sigma_j^i)^2}, & if\ x_j \leq m_{1,j}^i; \\ 0, & otherwise. \end{cases} \tag{S56}$$

$$\frac{\partial \underline{f}^{i,k}}{\partial m_{1,j}^i} = \frac{\partial \underline{f}^{i,k}}{\partial \underline{R}^i}\frac{\partial \underline{R}^i}{\partial m_{1,j}^i} = \begin{cases} \left(\underline{f}^{i,k}\right)^2 \times \dfrac{x_j - m_{1,j}^i}{(\sigma_j^i)^2}, & if\ x_j > \dfrac{m_{1,j}^i + m_{2,j}^i}{2}; \\ 0, & otherwise. \end{cases} \tag{S57}$$

$$\frac{\partial \overline{f}^{i,k}}{\partial m_{2,j}^i} = \frac{\partial \overline{f}^{i,k}}{\partial \overline{R}^i}\frac{\partial \overline{R}^i}{\partial m_{2,j}^i} = \begin{cases} \left(\overline{f}^{i,k}\right)^2 \times \dfrac{x_j - m_{2,j}^i}{(\sigma_j^i)^2}, & if\ x_j > m_{2,j}^i; \\ 0, & otherwise. \end{cases} \tag{S58}$$

$$\frac{\partial \underline{f}^{i,k}}{\partial m_{2,j}^i} = \frac{\partial \underline{f}^{i,k}}{\partial \underline{R}^i}\frac{\partial \underline{R}^i}{\partial m_{2,j}^i} = \begin{cases} \left(\underline{f}^{i,k}\right)^2 \times \dfrac{x_j - m_{2,j}^i}{(\sigma_j^i)^2}, & if\ x_j \leq \dfrac{m_{1,j}^i + m_{2,j}^i}{2}; \\ 0, & otherwise. \end{cases} \tag{S59}$$

$$\frac{\partial \overline{f}^{i,k}}{\partial \sigma_j^i} = \frac{\partial \overline{f}^{i,k}}{\partial \overline{R}^i}\frac{\partial \overline{R}^i}{\partial \sigma_j^i} = \begin{cases} \left(\overline{f}^{i,k}\right)^2 \times \dfrac{(x_j - m_{1,j}^i)^2}{(\sigma_j^i)^3}, & if\ x_j < m_{1,j}^i; \\ \left(\overline{f}^{i,k}\right)^2 \times \dfrac{(x_j - m_{2,j}^i)^2}{(\sigma_j^i)^3}, & if\ x_j > m_{2,j}^i; \\ 0, & otherwise. \end{cases} \tag{S60}$$

$$\frac{\partial \underline{f}^{i,k}}{\partial \sigma_j^i} = \frac{\partial \underline{f}^{i,k}}{\partial \underline{R}^i}\frac{\partial \underline{R}^i}{\partial \sigma_j^i} = \begin{cases} \left(\underline{f}^{i,k}\right)^2 \times \dfrac{(x_j - m_{2,j}^i)^2}{(\sigma_j^i)^3}, & if\ x_j \leq \dfrac{m_{1,j}^i + m_{2,j}^i}{2}; \\ \left(\underline{f}^{i,k}\right)^2 \times \dfrac{(x_j - m_{1,j}^i)^2}{(\sigma_j^i)^3}, & if\ x_j > \dfrac{m_{1,j}^i + m_{2,j}^i}{2}. \end{cases} \tag{S61}$$

$$\frac{\partial \overline{f}^{i,k}}{\partial m_j^k} = \frac{\partial \overline{f}^{i,k}}{\partial R^k}\frac{\partial R^k}{\partial m_j^k} = (\overline{f}^{i,k})^2 \times \frac{x_j - m_j^k}{(\sigma_j^k)^2} \tag{S62}$$

$$\frac{\partial \underline{f}^{i,k}}{\partial m_j^k} = \frac{\partial \underline{f}^{i,k}}{\partial R^k}\frac{\partial R^k}{\partial m_j^k} = (\underline{f}^{i,k})^2 \times \frac{x_j - m_j^k}{(\sigma_j^k)^2} \tag{S63}$$

$$\frac{\partial \overline{f}^{i,k}}{\partial \sigma_j^k} = \frac{\partial \overline{f}^{i,k}}{\partial R^k}\frac{\partial R^k}{\partial \sigma_j^k} = (\overline{f}^{i,k})^2 \times \frac{(x_j - m_j^k)^2}{(\sigma_j^k)^3} \tag{S64}$$

$$\frac{\partial \underline{f}^{i,k}}{\partial \sigma_j^k} = \frac{\partial \underline{f}^{i,k}}{\partial R^k}\frac{\partial R^k}{\partial \sigma_j^k} = (\underline{f}^{i,k})^2 \times \frac{(x_j - m_j^k)^2}{(\sigma_j^k)^3} \tag{S65}$$

## 2. SUPPLEMENT ON SIMULATIONS

Table S1: Performance comparison on models trained with clean chaotic time series in Example 1 (including results for each time step)

| Data | CNN-LSTM [11] | | | | | | CLF-IT2NN [25] | | | | | | SOIT2FNN | | | | **SOIT2FNN-MO** | |
|---|---|---|---|---|---|---|---|---|---|---|---|---|---|---|---|---|---|---|
| | SW | | PM | | MO | | SW | | PM | | MO | | SW | | PM | | | |
| | $y^1$-$y^3$ | Avg | $y^1$-$y^3$ | Avg | $y^1$-$y^3$ | Avg | $y^1$-$y^3$ | Avg | $y^1$-$y^3$ | Avg | $y^1$-$y^3$ | Avg | $y^1$-$y^3$ | Avg | $y^1$-$y^3$ | Avg | $y^1$-$y^3$ | Avg |
| Training | 0.006 | 0.009 | 0.003 | 0.008 | 0.006 | 0.006 | 0.034 | 0.043 | 0.034 | 0.036 | 0.026 | 0.037 | 0.038 | 0.055 | 0.038 | 0.035 | 0.021 | 0.027 |
| RMSE | 0.009 | | 0.007 | | 0.007 | | 0.044 | | 0.030 | | 0.038 | | 0.061 | | 0.030 | | 0.027 | |
| (Clean) | 0.011 | | 0.014 | | 0.006 | | 0.051 | | 0.044 | | 0.047 | | 0.066 | | 0.038 | | 0.033 | |
| Test | 0.006 | 0.009 | 0.003 | 0.008 | 0.006 | 0.007 | 0.035 | 0.043 | 0.035 | 0.039 | 0.032 | 0.040 | 0.038 | 0.057 | 0.038 | 0.036 | 0.024 | 0.031 |
| RMSE | 0.009 | | 0.007 | | 0.008 | | 0.043 | | 0.034 | | 0.038 | | 0.064 | | 0.031 | | 0.031 | |
| (Clean) | 0.011 | | 0.015 | | 0.006 | | 0.051 | | 0.048 | | 0.050 | | 0.069 | | 0.040 | | 0.036 | |
| Test | 0.141 | 0.149 | 0.141 | 0.149 | 0.142 | 0.149 | 0.165 | 0.201 | 0.165 | 0.169 | 0.179 | 0.169 | 0.204 | 0.191 | 0.204 | 0.176 | 0.169 | 0.165 |
| RMSE | 0.150 | | 0.150 | | 0.152 | | 0.204 | | 0.164 | | 0.180 | | 0.192 | | 0.159 | | 0.163 | |
| (std=10%) | 0.156 | | 0.156 | | 0.155 | | 0.234 | | 0.178 | | 0.148 | | 0.177 | | 0.163 | | 0.164 | |
| Test | 0.371 | 0.390 | 0.368 | 0.392 | 0.369 | 0.390 | 0.499 | 0.621 | 0.499 | 0.534 | 0.545 | 0.527 | 0.641 | 0.805 | 0.641 | 0.491 | 0.561 | 0.496 |
| RMSE | 0.393 | | 0.399 | | 0.392 | | 0.593 | | 0.572 | | 0.502 | | 0.816 | | 0.417 | | 0.510 | |
| (std=30%) | 0.406 | | 0.410 | | 0.409 | | 0.771 | | 0.531 | | 0.534 | | 0.959 | | 0.415 | | 0.416 | |
| Training | 0.603 | 0.903 | 0.319 | 0.858 | 0.605 | 0.722 | 3.315 | 4.892 | 3.115 | 3.132 | 2.189 | 3.298 | 3.612 | 5.413 | 3.612 | 3.446 | 2.196 | 2.709 |
| MPE | 0.941 | | 0.709 | | 0.887 | | 4.921 | | 2.925 | | 2.932 | | 6.016 | | 3.049 | | 2.886 | |
| (Clean) | 1.164 | | 1.546 | | 0.674 | | 6.440 | | 3.356 | | 4.773 | | 6.611 | | 3.676 | | 3.046 | |
| Test | 0.610 | 0.915 | 0.331 | 0.882 | 0.660 | 0.770 | 3.402 | 3.798 | 3.402 | 3.705 | 3.390 | 3.774 | 3.704 | 5.669 | 3.704 | 3.694 | 2.563 | 3.157 |
| MPE | 0.953 | | 0.716 | | 0.936 | | 3.807 | | 3.712 | | 3.720 | | 6.289 | | 3.136 | | 3.364 | |
| (Clean) | 1.184 | | 1.600 | | 0.713 | | 4.185 | | 4.001 | | 4.212 | | 7.015 | | 4.240 | | 3.545 | |
| Test | 15.03 | 16.18 | 15.04 | 15.90 | 15.16 | 16.19 | 17.35 | 21.42 | 17.35 | 19.13 | 18.90 | 19.00 | 20.82 | 19.88 | 20.82 | 18.81 | 18.65 | 18.61 |
| MPE | 16.39 | | 16.05 | | 16.50 | | 22.94 | | 17.01 | | 20.13 | | 20.01 | | 17.24 | | 18.73 | |
| (std=10%) | 17.12 | | 16.60 | | 16.90 | | 23.97 | | 23.03 | | 17.97 | | 18.80 | | 18.36 | | 18.44 | |
| Test | 69.17 | 77.44 | 69.01 | 75.04 | 70.10 | 74.88 | 64.31 | 70.25 | 64.31 | 68.32 | 72.25 | 68.39 | 84.36 | 86.90 | 84.36 | 67.46 | 79.00 | 67.20 |
| MPE | 80.34 | | 77.50 | | 76.71 | | 70.22 | | 73.25 | | 66.33 | | 86.25 | | 58.25 | | 64.48 | |
| (std=30%) | 82.81 | | 78.60 | | 77.83 | | 76.22 | | 67.40 | | 66.59 | | 90.09 | | 59.78 | | 58.13 | |
| Rule | — | | — | | — | | 2 | | 2,2,2 | | 2 | | 1 | | 1,2,2 | | 2 | |

Table S2: Performance comparison on models trained with noisy chaotic time series (std = 10%) in Example 1 (including results for each time step)

| Data | CNN-LSTM [11] | | | | | | CLF-IT2NN [25] | | | | | | SOIT2FNN | | | | **SOIT2FNN-MO** | |
|---|---|---|---|---|---|---|---|---|---|---|---|---|---|---|---|---|---|---|
| | SW | | PM | | MO | | SW | | PM | | MO | | SW | | PM | | | |
| | $y^1$-$y^3$ | Avg | $y^1$-$y^3$ | Avg | $y^1$-$y^3$ | Avg | $y^1$-$y^3$ | Avg | $y^1$-$y^3$ | Avg | $y^1$-$y^3$ | Avg | $y^1$-$y^3$ | Avg | $y^1$-$y^3$ | Avg | $y^1$-$y^3$ | Avg |
| Training | 0.129 | 0.137 | 0.121 | 0.124 | 0.119 | 0.122 | 0.123 | 0.134 | 0.123 | 0.131 | 0.128 | 0.133 | 0.127 | 0.128 | 0.125 | 0.128 | 0.125 | 0.130 |
| RMSE | 0.138 | | 0.124 | | 0.124 | | 0.135 | | 0.131 | | 0.131 | | 0.128 | | 0.127 | | 0.131 | |
| (std=10%) | 0.144 | | 0.126 | | 0.124 | | 0.144 | | 0.139 | | 0.140 | | 0.131 | | 0.132 | | 0.133 | |
| Test | 0.048 | 0.059 | 0.038 | 0.046 | 0.048 | 0.049 | 0.051 | 0.062 | 0.051 | 0.057 | 0.49 | 0.054 | 0.050 | 0.054 | 0.043 | 0.049 | 0.043 | 0.046 |
| RMSE | 0.061 | | 0.048 | | 0.052 | | 0.063 | | 0.055 | | 0.055 | | 0.055 | | 0.047 | | 0.048 | |
| (Clean) | 0.068 | | 0.052 | | 0.047 | | 0.072 | | 0.065 | | 0.058 | | 0.057 | | 0.058 | | 0.048 | |
| Test | 0.133 | 0.142 | 0.131 | 0.140 | 0.135 | 0.140 | 0.157 | 0.191 | 0.157 | 0.173 | 0.140 | 0.152 | 0.134 | 0.142 | 0.131 | 0.139 | 0.131 | 0.140 |
| RMSE | 0.143 | | 0.141 | | 0.142 | | 0.188 | | 0.180 | | 0.151 | | 0.144 | | 0.139 | | 0.141 | |
| (std=10%) | 0.150 | | 0.148 | | 0.145 | | 0.228 | | 0.182 | | 0.165 | | 0.148 | | 0.148 | | 0.148 | |
| Test | 0.358 | 0.378 | 0.348 | 0.357 | 0.350 | 0.357 | 0.305 | 0.401 | 0.305 | 0.388 | 0.363 | 0.372 | 0.342 | 0.353 | 0.335 | 0.349 | 0.335 | 0.348 |
| RMSE | 0.384 | | 0.358 | | 0.360 | | 0.412 | | 0.391 | | 0.377 | | 0.357 | | 0.353 | | 0.353 | |
| (std=30%) | 0.392 | | 0.365 | | 0.360 | | 0.486 | | 0.468 | | 0.376 | | 0.360 | | 0.359 | | 0.357 | |
| Training | 13.94 | 14.46 | 13.95 | 14.40 | 13.81 | 14.22 | 13.50 | 14.98 | 13.50 | 14.00 | 14.66 | 15.22 | 14.67 | 14.90 | 14.32 | 14.67 | 14.32 | 14.79 |
| MPE | 14.52 | | 14.29 | | 14.59 | | 15.09 | | 13.91 | | 14.98 | | 14.87 | | 14.48 | | 14.92 | |
| (std=10%) | 14.92 | | 14.95 | | 14.25 | | 16.35 | | 14.59 | | 16.02 | | 15.16 | | 15.22 | | 15.15 | |
| Test | 4.737 | 5.683 | 4.122 | 4.998 | 4.805 | 5.119 | 5.136 | 5.547 | 5.136 | 5.216 | 4.887 | 5.219 | 5.048 | 5.385 | 4.555 | 5.085 | 4.555 | 4.901 |
| MPE | 5.792 | | 4.977 | | 5.573 | | 5.662 | | 5.257 | | 5.154 | | 5.479 | | 4.951 | | 5.123 | |
| (Clean) | 6.520 | | 5.894 | | 4.980 | | 5.843 | | 5.255 | | 5.616 | | 5.626 | | 5.748 | | 5.027 | |
| Test | 13.70 | 14.59 | 14.62 | 15.91 | 15.16 | 15.92 | 15.62 | 18.39 | 15.62 | 16.55 | 15.02 | 16.32 | 14.76 | 15.92 | 14.35 | 15.51 | 14.35 | 15.40 |
| MPE | 14.64 | | 15.78 | | 16.18 | | 17.01 | | 16.90 | | 16.44 | | 16.14 | | 15.22 | | 15.58 | |
| (std=10%) | 15.43 | | 17.34 | | 16.43 | | 22.54 | | 17.13 | | 17.50 | | 16.86 | | 16.96 | | 16.28 | |
| Test | 81.32 | 83.84 | 78.82 | 81.42 | 73.68 | 76.09 | 38.94 | 51.17 | 38.94 | 46.08 | 40.33 | 45.22 | 38.56 | 41.51 | 42.28 | 41.23 | 40.24 | 41.38 |
| MPE | 82.68 | | 79.84 | | 77.35 | | 52.24 | | 45.51 | | 47.87 | | 41.34 | | 40.11 | | 41.61 | |
| (std=30%) | 87.52 | | 85.59 | | 77.23 | | 62.33 | | 53.79 | | 47.46 | | 44.64 | | 41.29 | | 42.28 | |
| Rule | —— | | —— | | —— | | 2 | | 2,2,3 | | 3 | | 2 | | 2,2,2 | | 2 | |

Table S3: Performance comparison on models trained with noisy chaotic time series (std=30%) in Example 1 (including results for each time step)

| Data | CNN-LSTM [11] | | | | | | CLF-IT2NN [25] | | | | | | SOIT2FNN | | | | **SOIT2FNN-MO** | |
|---|---|---|---|---|---|---|---|---|---|---|---|---|---|---|---|---|---|---|
| | SW | | PM | | MO | | SW | | PM | | MO | | SW | | PM | | | |
| | $y^1$-$y^3$ | Avg | $y^1$-$y^3$ | Avg | $y^1$-$y^3$ | Avg | $y^1$-$y^3$ | Avg | $y^1$-$y^3$ | Avg | $y^1$-$y^3$ | Avg | $y^1$-$y^3$ | Avg | $y^1$-$y^3$ | Avg | $y^1$-$y^3$ | Avg |
| Training | 0.068 | 0.112 | 0.114 | 0.070 | 0.105 | 0.108 | 0.351 | 0.368 | 0.365 | 0.397 | 0.339 | 0.355 | 0.347 | 0.359 | 0.333 | 0.415 | 0.335 | 0.339 |
| RMSE | 0.113 | | 0.065 | | 0.103 | | 0.367 | | 0.408 | | 0.352 | | 0.359 | | 0.432 | | 0.339 | |
| (std=30%) | 0.155 | | 0.031 | | 0.116 | | 0.386 | | 0.418 | | 0.374 | | 0.370 | | 0.479 | | 0.344 | |
| Test | 0.340 | 0.338 | 0.290 | 0.314 | 0.310 | 0.323 | 0.167 | 0.185 | 0.211 | 0.209 | 0.179 | 0.177 | 0.145 | 0.170 | 0.136 | 0.240 | 0.153 | 0.157 |
| RMSE | 0.312 | | 0.259 | | 0.341 | | 0.174 | | 0.207 | | 0.165 | | 0.169 | | 0.269 | | 0.142 | |
| (Clean) | 0.363 | | 0.394 | | 0.320 | | 0.214 | | 0.209 | | 0.187 | | 0.197 | | 0.314 | | 0.176 | |
| Test | 0.354 | 0.379 | 0.304 | 0.341 | 0.333 | 0.335 | 0.230 | 0.257 | 0.199 | 0.225 | 0.180 | 0.192 | 0.178 | 0.205 | 0.174 | 0.266 | 0.186 | 0.187 |
| RMSE | 0.369 | | 0.297 | | 0.336 | | 0.264 | | 0.218 | | 0.195 | | 0.205 | | 0.290 | | 0.177 | |
| (std=10%) | 0.414 | | 0.423 | | 0.335 | | 0.277 | | 0.258 | | 0.201 | | 0.231 | | 0.333 | | 0.199 | |
| Test | 0.589 | 0.638 | 0.577 | 0.589 | 0.584 | 0.557 | 0.388 | 0.412 | 0.394 | 0.440 | 0.355 | 0.386 | 0.346 | 0.363 | 0.352 | 0.411 | 0.351 | 0.358 |
| RMSE | 0.629 | | 0.550 | | 0.532 | | 0.421 | | 0.425 | | 0.376 | | 0.369 | | 0.423 | | 0.36 | |
| (std=30%) | 0.697 | | 0.640 | | 0.556 | | 0.427 | | 0.501 | | 0.427 | | 0.373 | | 0.457 | | 0.362 | |
| Training | 8.971 | 16.33 | 14.33 | 9.274 | 16.25 | 16.25 | 28.45 | 30.00 | 26.35 | 30.35 | 24.21 | 26.28 | 27.94 | 29.73 | 30.45 | 34.85 | 26.25 | 25.87 |
| MPE | 13.73 | | 7.822 | | 14.03 | | 29.39 | | 32.08 | | 26.23 | | 30.40 | | 35.77 | | 26.03 | |
| (std=30%) | 26.27 | | 5.669 | | 18.46 | | 32.16 | | 32.32 | | 28.40 | | 30.87 | | 38.33 | | 25.33 | |
| Test | 31.46 | 32.80 | 28.12 | 31.43 | 32.18 | 32.05 | 16.27 | 18.88 | 18.35 | 19.73 | 15.85 | 16.68 | 17.61 | 21.54 | 15.55 | 22.83 | 16.48 | 15.67 |
| MPE | 30.83 | | 28.16 | | 30.74 | | 18.09 | | 19.86 | | 15.07 | | 21.34 | | 25.04 | | 13.64 | |
| (Clean) | 36.10 | | 38.03 | | 33.23 | | 22.28 | | 20.98 | | 19.12 | | 25.66 | | 27.91 | | 16.89 | |
| Test | 34.90 | 38.42 | 31.55 | 36.61 | 35.63 | 35.12 | 23.35 | 26.34 | 20.45 | 23.50 | 21.35 | 23.04 | 21.74 | 27.44 | 22.67 | 27.09 | 21.65 | 21.98 |
| MPE | 38.69 | | 33.67 | | 34.81 | | 27.04 | | 22.82 | | 22.98 | | 29.24 | | 27.00 | | 20.81 | |
| (std=10%) | 41.66 | | 44.60 | | 34.93 | | 28.63 | | 27.23 | | 24.79 | | 31.34 | | 31.61 | | 23.48 | |
| Test | 108.50 | 115.47 | 109.9 | 101.61 | 96.23 | 96.06 | 53.51 | 57.21 | 52.25 | 59.25 | 42.22 | 48.63 | 49.72 | 55.04 | 45.18 | 46.24 | 49.76 | 44.48 |
| MPE | 112.97 | | 98.04 | | 77.46 | | 58.35 | | 58.67 | | 47.35 | | 56.24 | | 47.87 | | 43.28 | |
| (std=30%) | 124.93 | | 96.91 | | 114.48 | | 59.77 | | 66.83 | | 56.32 | | 59.15 | | 45.67 | | 40.39 | |
| Rule | —— | | —— | | —— | | 3 | | 2,3,3 | | 5 | | 2 | | 2,3,3 | | 5 | |

Table S4: Performance comparison on microgrid unmet power in Example 2 (including results for each time step)

| Data | CNN-LSTM [11] | | | | | | CLF-IT2NN [25] | | | | | | SOIT2FNN | | | | **SOIT2FNN-MO** | |
|---|---|---|---|---|---|---|---|---|---|---|---|---|---|---|---|---|---|---|
| | SW | | PM | | MO | | SW | | PM | | MO | | SW | | PM | | | |
| | $y^1$-$y^3$ | Avg | $y^1$-$y^3$ | Avg | $y^1$-$y^3$ | Avg | $y^1$-$y^3$ | Avg | $y^1$-$y^3$ | Avg | $y^1$-$y^3$ | Avg | $y^1$-$y^3$ | Avg | $y^1$-$y^3$ | Avg | $y^1$-$y^3$ | Avg |
| Training | 0.044 | 0.070 | 0.044 | 0.061 | 0.037 | 0.052 | 0.082 | 0.111 | 0.075 | 0.097 | 0.079 | 0.103 | 0.057 | 0.091 | 0.057 | 0.090 | 0.054 | 0.084 |
| RMSE | 0.071 | | 0.063 | | 0.053 | | 0.120 | | 0.099 | | 0.102 | | 0.095 | | 0.094 | | 0.088 | |
| (Clean) | 0.096 | | 0.076 | | 0.065 | | 0.130 | | 0.117 | | 0.128 | | 0.121 | | 0.119 | | 0.111 | |
| Test | 0.045 | 0.072 | 0.046 | 0.069 | 0.050 | 0.074 | 0.082 | 0.109 | 0.069 | 0.087 | 0.074 | 0.089 | 0.055 | 0.088 | 0.055 | 0.087 | 0.052 | 0.080 |
| RMSE | 0.073 | | 0.072 | | 0.077 | | 0.099 | | 0.092 | | 0.088 | | 0.091 | | 0.090 | | 0.084 | |
| (Clean) | 0.098 | | 0.091 | | 0.095 | | 0.146 | | 0.100 | | 0.105 | | 0.117 | | 0.115 | | 0.105 | |
| Test | 0.114 | 0.152 | 0.107 | 0.155 | 0.109 | 0.157 | 0.129 | 0.162 | 0.110 | 0.145 | 0.093 | 0.123 | 0.095 | 0.120 | 0.095 | 0.120 | 0.093 | 0.116 |
| RMSE | 0.157 | | 0.173 | | 0.161 | | 0.160 | | 0.138 | | 0.114 | | 0.125 | | 0.124 | | 0.120 | |
| (std=10%) | 0.186 | | 0.187 | | 0.200 | | 0.197 | | 0.187 | | 0.162 | | 0.140 | | 0.140 | | 0.136 | |
| Test | 0.282 | 0.327 | 0.279 | 0.326 | 0.246 | 0.324 | 0.245 | 0.282 | 0.217 | 0.269 | 0.224 | 0.264 | 0.226 | 0.249 | 0.226 | 0.248 | 0.228 | 0.249 |
| RMSE | 0.328 | | 0.346 | | 0.328 | | 0.288 | | 0.263 | | 0.259 | | 0.253 | | 0.251 | | 0.252 | |
| (std=30%) | 0.372 | | 0.354 | | 0.399 | | 0.313 | | 0.327 | | 0.309 | | 0.269 | | 0.267 | | 0.266 | |
| Training | 3.308 | 6.083 | 3.212 | 5.013 | 2.943 | 4.215 | 6.403 | 8.203 | 7.882 | 8.154 | 6.210 | 7.945 | 4.668 | 8.198 | 4.668 | 8.116 | 4.291 | 7.358 |
| MPE | 6.110 | | 5.203 | | 4.245 | | 7.615 | | 8.165 | | 7.825 | | 8.520 | | 8.445 | | 7.610 | |
| (Clean) | 8.831 | | 6.624 | | 5.457 | | 10.59 | | 8.415 | | 9.801 | | 11.40 | | 11.23 | | 10.17 | |
| Test | 3.503 | 6.367 | 3.468 | 5.831 | 3.864 | 6.014 | 8.044 | 8.355 | 7.035 | 8.079 | 6.841 | 8.152 | 4.650 | 8.206 | 4.650 | 8.097 | 4.257 | 7.258 |
| MPE | 6.390 | | 6.131 | | 6.188 | | 8.357 | | 7.968 | | 8.077 | | 8.503 | | 8.409 | | 7.513 | |
| (Clean) | 9.206 | | 7.895 | | 7.988 | | 8.664 | | 9.234 | | 9.538 | | 11.47 | | 11.23 | | 10.00 | |
| Test | 10.68 | 14.68 | 10.44 | 15.05 | 10.60 | 15.44 | 11.17 | 14.76 | 10.21 | 13.33 | 10.03 | 12.05 | 9.213 | 11.80 | 9.213 | 11.80 | 9.008 | 11.38 |
| MPE | 14.98 | | 16.22 | | 15.81 | | 14.65 | | 12.43 | | 11.82 | | 12.16 | | 12.12 | | 11.69 | |
| (std=10%) | 18.38 | | 18.49 | | 19.92 | | 18.46 | | 17.35 | | 14.30 | | 14.02 | | 14.05 | | 13.44 | |
| Test | 28.98 | 34.88 | 28.99 | 34.34 | 26.76 | 34.56 | 25.34 | 31.42 | 22.57 | 27.88 | 24.48 | 28.31 | 24.12 | 26.48 | 24.12 | 26.60 | 23.75 | 26.27 |
| MPE | 34.79 | | 35.98 | | 34.58 | | 29.09 | | 28.10 | | 26.83 | | 26.93 | | 27.09 | | 26.78 | |
| (std=30%) | 40.88 | | 38.05 | | 42.34 | | 39.83 | | 32.97 | | 33.62 | | 28.39 | | 28.58 | | 28.29 | |
| Rule | — | | — | | — | | 1 | | 1,1,2 | | 3 | | 1 | | 1,1,1 | | 3 | |

Table S5: Performance comparison on the electricity price in Example 2 (including results for each time step)

| Data | CNN-LSTM [11] | | | | | | CLF-IT2NN [25] | | | | | | SOIT2FNN | | | | **SOIT2FNN-MO** | |
|---|---|---|---|---|---|---|---|---|---|---|---|---|---|---|---|---|---|---|
| | SW | | PM | | MO | | SW | | PM | | MO | | SW | | PM | | | |
| | $y^1$-$y^3$ | Avg | $y^1$-$y^3$ | Avg | $y^1$-$y^3$ | Avg | $y^1$-$y^3$ | Avg | $y^1$-$y^3$ | Avg | $y^1$-$y^3$ | Avg | $y^1$-$y^3$ | Avg | $y^1$-$y^3$ | Avg | $y^1$-$y^3$ | Avg |
| Training | 0.036 | 0.059 | 0.037 | 0.051 | 0.036 | 0.050 | 0.057 | 0.088 | 0.057 | 0.080 | 0.054 | 0.082 | 0.050 | 0.077 | 0.050 | 0.077 | 0.048 | 0.074 |
| RMSE | 0.058 | | 0.052 | | 0.052 | | 0.073 | | 0.082 | | 0.076 | | 0.080 | | 0.079 | | 0.076 | |
| (Clean) | 0.081 | | 0.063 | | 0.064 | | 0.134 | | 0.101 | | 0.116 | | 0.101 | | 0.101 | | 0.096 | |
| Test | 0.039 | 0.061 | 0.038 | 0.056 | 0.040 | 0.058 | 0.052 | 0.080 | 0.052 | 0.078 | 0.049 | 0.076 | 0.045 | 0.072 | 0.045 | 0.071 | 0.044 | 0.068 |
| RMSE | 0.062 | | 0.058 | | 0.060 | | 0.092 | | 0.081 | | 0.073 | | 0.074 | | 0.074 | | 0.071 | |
| (Clean) | 0.083 | | 0.071 | | 0.074 | | 0.96 | | 0.101 | | 0.106 | | 0.096 | | 0.095 | | 0.090 | |
| Test | 0.063 | 0.085 | 0.064 | 0.086 | 0.064 | 0.086 | 0.079 | 0.116 | 0.079 | 0.102 | 0.084 | 0.098 | 0.066 | 0.089 | 0.066 | 0.087 | 0.065 | 0.086 |
| RMSE | 0.086 | | 0.086 | | 0.087 | | 0.099 | | 0.095 | | 0.098 | | 0.091 | | 0.089 | | 0.088 | |
| (std=10%) | 0.107 | | 0.107 | | 0.108 | | 0.170 | | 0.132 | | 0.112 | | 0.110 | | 0.107 | | 0.105 | |
| Test | 0.159 | 0.175 | 0.160 | 0.193 | 0.164 | 0.205 | 0.135 | 0.189 | 0.135 | 0.172 | 0.140 | 0.173 | 0.149 | 0.163 | 0.149 | 0.160 | 0.146 | 0.160 |
| RMSE | 0.174 | | 0.195 | | 0.207 | | 0.184 | | 0.168 | | 0.182 | | 0.163 | | 0.160 | | 0.160 | |
| (std=30%) | 0.193 | | 0.224 | | 0.246 | | 0.248 | | 0.213 | | 0.197 | | 0.178 | | 0.172 | | 0.175 | |
| Training | 5.434 | 9.094 | 5.551 | 8.125 | 5.776 | 8.283 | 7.772 | 12.80 | 7.772 | 11.89 | 7.351 | 12.32 | 6.645 | 11.54 | 6.645 | 11.32 | 6.561 | 10.90 |
| MPE | 9.103 | | 8.342 | | 8.604 | | 11.56 | | 10.98 | | 11.25 | | 11.81 | | 11.59 | | 11.26 | |
| (Clean) | 12.74 | | 10.48 | | 10.47 | | 19.07 | | 16.92 | | 18.36 | | 16.17 | | 15.74 | | 14.88 | |
| Test | 5.942 | 10.18 | 5.806 | 9.256 | 6.431 | 9.748 | 6.983 | 12.00 | 6.983 | 11.71 | 7.021 | 11.73 | 6.582 | 11.73 | 6.582 | 11.47 | 6.497 | 11.22 |
| MPE | 10.25 | | 9.679 | | 10.06 | | 13.81 | | 12.32 | | 10.28 | | 11.98 | | 11.73 | | 11.53 | |
| (Clean) | 14.34 | | 12.28 | | 12.75 | | 15.21 | | 15.83 | | 17.89 | | 16.62 | | 16.11 | | 15.63 | |
| Test | 12.16 | 15.82 | 12.37 | 15.85 | 12.42 | 16.19 | 12.73 | 20.02 | 12.73 | 18.44 | 15.35 | 17.56 | 12.78 | 17.09 | 12.78 | 16.47 | 12.60 | 16.43 |
| MPE | 15.70 | | 15.72 | | 16.21 | | 19.97 | | 18.34 | | 17.42 | | 17.24 | | 16.56 | | 16.61 | |
| (std=10%) | 19.60 | | 19.46 | | 19.94 | | 27.36 | | 24.25 | | 19.91 | | 21.25 | | 20.08 | | 20.08 | |
| Test | 39.93 | 40.63 | 39.92 | 44.79 | 41.25 | 48.08 | 27.58 | 42.52 | 27.58 | 37.87 | 29.76 | 38.25 | 30.79 | 37.87 | 30.79 | 37.20 | 29.41 | 36.67 |
| MPE | 39.92 | | 44.68 | | 47.97 | | 43.09 | | 41.36 | | 41.35 | | 41.04 | | 39.76 | | 39.91 | |
| (std=30%) | 42.05 | | 49.77 | | 55.03 | | 56.89 | | 44.67 | | 43.64 | | 41.79 | | 41.04 | | 40.69 | |
| Rule | —— | | —— | | —— | | 1 | | 1,1,2 | | 3 | | 1 | | 1,1,1 | | 2 | |

Table S6: Performance comparison on clean training/test datasets regarding different clustering numbers (including results for each time step)

| Data | No. = 5 | | No. = 10 | | No. = 15 | | No. = 20 | | No. = 25 | | No. = 30 | |
|---|---|---|---|---|---|---|---|---|---|---|---|---|
| | $y^1$-$y^3$ | Avg | $y^1$-$y^3$ | Avg | $y^1$-$y^3$ | Avg | $y^1$-$y^3$ | Avg | $y^1$-$y^3$ | Avg | $y^1$-$y^3$ | Avg |
| Training | 0.0213 | 0.0274 | 0.0224 | 0.0277 | 0.0192 | 0.0250 | 0.0198 | 0.0236 | 0.0204 | 0.0264 | 0.0203 | 0.0230 |
| RMSE | 0.0274 | | 0.0306 | | 0.0228 | | 0.0228 | | 0.0265 | | 0.0224 | |
| (Chaotic) | 0.0334 | | 0.0301 | | 0.0329 | | 0.0283 | | 0.0322 | | 0.0264 | |
| Training | 2.1962 | 2.7094 | 2.4307 | 3.0814 | 2.0508 | 2.6263 | 1.9870 | 2.4072 | 2.1455 | 2.8162 | 2.1010 | 2.4164 |
| MPE | 2.8857 | | 3.4367 | | 2.3792 | | 2.3633 | | 2.8160 | | 2.3825 | |
| (Chaotic) | 3.0463 | | 3.3767 | | 3.4417 | | 2.8712 | | 3.4872 | | 2.7656 | |
| Test | 0.0242 | 0.0305 | 0.0268 | 0.0346 | 0.0211 | 0.0307 | 0.0214 | 0.0293 | 0.0231 | 0.0342 | 0.0227 | 0.0286 |
| RMSE | 0.0312 | | 0.0369 | | 0.0299 | | 0.0299 | | 0.0354 | | 0.0276 | |
| (Chaotic) | 0.0360 | | 0.0402 | | 0.0411 | | 0.0367 | | 0.0440 | | 0.0356 | |
| Test | 2.5625 | 3.1571 | 2.9790 | 3.8622 | 2.2431 | 3.1090 | 2.2192 | 2.9386 | 2.4485 | 3.5890 | 2.3838 | 2.9345 |
| MPE | 3.3641 | | 4.1745 | | 2.8741 | | 2.8950 | | 3.6732 | | 2.8115 | |
| (Chaotic) | 3.5446 | | 4.4331 | | 4.2097 | | 3.7017 | | 4.6452 | | 3.6081 | |
| Training | 0.0544 | 0.0843 | 0.0544 | 0.0846 | 0.0531 | 0.0831 | 0.0533 | 0.0834 | 0.0535 | 0.0823 | 0.0281 | 0.0806 |
| RMSE | 0.0876 | | 0.0879 | | 0.0861 | | 0.0864 | | 0.0856 | | 0.0700 | |
| (power) | 0.1109 | | 0.1113 | | 0.1100 | | 0.1105 | | 0.1079 | | 0.1107 | |
| Training | 4.2908 | 7.3576 | 4.2961 | 7.3615 | 4.1641 | 7.1896 | 4.1779 | 7.2133 | 4.2301 | 7.1736 | 4.2122 | 7.0331 |
| MPE | 7.6103 | | 7.6340 | | 7.4007 | | 7.4152 | | 7.4380 | | 7.2718 | |
| (power) | 10.172 | | 10.154 | | 10.004 | | 10.047 | | 9.8528 | | 9.6151 | |
| Test | 0.0522 | 0.0803 | 0.0523 | 0.0810 | 0.0503 | 0.0787 | 0.0505 | 0.0790 | 0.0509 | 0.0783 | 0.0509 | 0.0774 |
| RMSE | 0.0837 | | 0.0845 | | 0.0817 | | 0.0818 | | 0.0812 | | 0.0803 | |
| (power) | 0.1049 | | 0.1062 | | 0.1043 | | 0.1047 | | 0.1028 | | 0.1011 | |
| Test | 4.2568 | 7.2580 | 4.3070 | 7.3772 | 4.1640 | 7.2036 | 4.1973 | 7.2185 | 4.1686 | 7.0742 | 4.2083 | 7.0293 |
| MPE | 7.5134 | | 7.6535 | | 7.4223 | | 7.4204 | | 7.3084 | | 7.2520 | |
| (power) | 10.004 | | 10.171 | | 10.024 | | 10.038 | | 9.7455 | | 9.6277 | |
| Training | 0.0499 | 0.0772 | 0.0484 | 0.0736 | 0.0495 | 0.0762 | 0.0469 | 0.0701 | 0.0484 | 0.0737 | 0.0524 | 0.0805 |
| RMSE | 0.0799 | | 0.0762 | | 0.0789 | | 0.0725 | | 0.0763 | | 0.0826 | |
| (price) | 0.1018 | | 0.0961 | | 0.1002 | | 0.0910 | | 0.0965 | | 0.1065 | |
| Training | 6.6753 | 11.606 | 6.5607 | 10.900 | 6.5872 | 11.178 | 6.5900 | 10.682 | 6.5521 | 10.8631 | 7.6863 | 12.453 |
| MPE | 11.849 | | 11.257 | | 11.509 | | 11.047 | | 11.1950 | | 12.427 | |
| (price) | 16.294 | | 14.882 | | 15.444 | | 14.407 | | 14.8422 | | 17.246 | |
| Test | 0.0455 | 0.0719 | 0.0439 | 0.0683 | 0.0200 | 0.0710 | 0.0436 | 0.0663 | 0.0438 | 0.0684 | 0.0480 | 0.0741 |
| RMSE | 0.0745 | | 0.0707 | | 0.0541 | | 0.0687 | | 0.0707 | | 0.0761 | |
| (price) | 0.0958 | | 0.0904 | | 0.0892 | | 0.0865 | | 0.0907 | | 0.0983 | |
| Test | 6.5861 | 11.680 | 6.4973 | 11.220 | 6.5114 | 11.320 | 6.6343 | 11.041 | 6.4226 | 11.084 | 7.6119 | 12.450 |
| MPE | 11.940 | | 11.530 | | 11.664 | | 11.410 | | 11.373 | | 12.403 | |
| (price) | 16.574 | | 15.634 | | 15.784 | | 15.079 | | 15.455 | | 17.335 | |
| Rule (Chaotic) | | 2 | | 2 | | 2 | | 2 | | 2 | | 2 |
| Rule (power) | | 3 | | 3 | | 3 | | 3 | | 3 | | 4 |
| Rule (price) | | 1 | | 2 | | 1 | | 3 | | 2 | | 1 |

Table S7: RMSE evaluation for each modified/added layer (including results for each time step)

| Data | SIT2FNN [52] | | Our | | S+L4 | | S+L6 | | S+L9 | |
|---|---|---|---|---|---|---|---|---|---|---|
| | $y^1$-$y^3$ | Avg | $y^1$-$y^3$ | Avg | $y^1$-$y^3$ | Avg | $y^1$-$y^3$ | Avg | $y^1$-$y^3$ | Avg |
| Chaotic | 0.040 | 0.041 | 0.024 | 0.031 | 0.037 | 0.043 | 0.024 | 0.031 | 0.033 | 0.039 |
| Clean | 0.040 | | 0.031 | | 0.045 | | 0.031 | | 0.042 | |
| RMSE | 0.043 | | 0.040 | | 0.046 | | 0.039 | | 0.042 | |
| Chaotic | 0.162 | 0.173 | 0.169 | 0.165 | 0.164 | 0.169 | 0.186 | 0.176 | 0.163 | 0.173 |
| std=10% | 0.174 | | 0.163 | | 0.165 | | 0.167 | | 0.175 | |
| RMSE | 0.181 | | 0.164 | | 0.179 | | 0.176 | | 0.182 | |
| Chaotic | 0.578 | 0.529 | 0.561 | 0.496 | 0.474 | 0.492 | 0.520 | 0.503 | 0.491 | 0.490 |
| std=30% | 0.514 | | 0.510 | | 0.497 | | 0.475 | | 0.480 | |
| RMSE | 0.494 | | 0.416 | | 0.505 | | 0.514 | | 0.499 | |
| Power | 0.057 | 0.086 | 0.052 | 0.080 | 0.054 | 0.083 | 0.057 | 0.089 | 0.057 | 0.086 |
| Clean | 0.087 | | 0.084 | | 0.087 | | 0.092 | | 0.087 | |
| RMSE | 0.113 | | 0.105 | | 0.109 | | 0.117 | | 0.113 | |
| Power | 0.096 | 0.120 | 0.093 | 0.116 | 0.095 | 0.118 | 0.096 | 0.117 | 0.097 | 0.117 |
| std=10% | 0.125 | | 0.120 | | 0.120 | | 0.119 | | 0.119 | |
| RMSE | 0.140 | | 0.136 | | 0.138 | | 0.137 | | 0.136 | |
| Power | 0.255 | 0.265 | 0.228 | 0.249 | 0.240 | 0.253 | 0.232 | 0.250 | 0.242 | 0.250 |
| std=30% | 0.263 | | 0.252 | | 0.251 | | 0.254 | | 0.248 | |
| RMSE | 0.277 | | 0.266 | | 0.268 | | 0.265 | | 0.261 | |
| Price | 0.057 | 0.083 | 0.044 | 0.068 | 0.046 | 0.073 | 0.048 | 0.074 | 0.047 | 0.073 |
| Clean | 0.084 | | 0.071 | | 0.075 | | 0.077 | | 0.074 | |
| RMSE | 0.107 | | 0.090 | | 0.097 | | 0.098 | | 0.097 | |
| Price | 0.068 | 0.092 | 0.065 | 0.086 | 0.065 | 0.088 | 0.066 | 0.088 | 0.065 | 0.088 |
| std=10% | 0.090 | | 0.088 | | 0.089 | | 0.090 | | 0.089 | |
| RMSE | 0.118 | | 0.105 | | 0.109 | | 0.108 | | 0.109 | |
| Price | 0.156 | 0.169 | 0.146 | 0.160 | 0.151 | 0.164 | 0.148 | 0.161 | 0.147 | 0.160 |
| std=30% | 0.168 | | 0.160 | | 0.163 | | 0.162 | | 0.159 | |
| RMSE | 0.184 | | 0.175 | | 0.179 | | 0.174 | | 0.175 | |
| Rule (Chaotic) | | 3 | | 2 | | 3 | | 2 | | 3 |
| Rule (power) | | 2 | | 3 | | 3 | | 1 | | 2 |
| Rule (price) | | 2 | | 2 | | 2 | | 2 | | 2 |

Table S8: MPE evaluation for each modified/added layer (including results for each time step)

| Data | SIT2FNN [52] | | Our | | S+L4 | | S+L6 | | S+L9 | |
|---|---|---|---|---|---|---|---|---|---|---|
| | $y^1$-$y^3$ | Avg | $y^1$-$y^3$ | Avg | $y^1$-$y^3$ | Avg | $y^1$-$y^3$ | Avg | $y^1$-$y^3$ | Avg |
| Chaotic | 4.448 | 4.548 | 2.563 | 3.157 | 3.679 | 4.216 | 2.503 | 3.315 | 3.754 | 4.292 |
| Clean | 4.491 | | 3.364 | | 4.561 | | 3.307 | | 4.658 | |
| MPE | 4.706 | | 3.545 | | 4.407 | | 4.136 | | 4.464 | |
| Chaotic | 20.50 | 20.27 | 18.65 | 18.61 | 17.94 | 18.66 | 18.67 | 19.79 | 18.70 | 20.00 |
| std=10% | 19.73 | | 18.73 | | 18.26 | | 20.03 | | 19.98 | |
| MPE | 20.57 | | 18.44 | | 19.79 | | 20.66 | | 21.33 | |
| Chaotic | 73.79 | 73.28 | 79.00 | 67.20 | 64.87 | 69.38 | 80.21 | 69.32 | 68.84 | 66.63 |
| std=30% | 64.82 | | 64.48 | | 73.25 | | 67.13 | | 58.01 | |
| MPE | 81.22 | | 58.13 | | 70.02 | | 60.62 | | 73.03 | |
| Power | 5.158 | 8.152 | 4.257 | 7.258 | 4.556 | 7.758 | 4.584 | 8.050 | 5.137 | 8.154 |
| Clean | 8.339 | | 7.513 | | 8.188 | | 8.313 | | 8.336 | |
| MPE | 10.96 | | 10.00 | | 10.53 | | 11.25 | | 10.99 | |
| Power | 10.10 | 12.65 | 9.008 | 11.38 | 10.20 | 12.54 | 9.386 | 11.58 | 9.560 | 11.69 |
| std=10% | 12.98 | | 11.69 | | 12.74 | | 11.70 | | 11.78 | |
| MPE | 14.88 | | 13.44 | | 14.67 | | 13.67 | | 13.72 | |
| Power | 27.16 | 28.26 | 23.75 | 26.27 | 24.81 | 26.19 | 23.41 | 25.65 | 24.63 | 25.87 |
| std=30% | 27.95 | | 26.78 | | 26.22 | | 26.28 | | 25.90 | |
| MPE | 29.66 | | 28.29 | | 27.55 | | 27.26 | | 27.07 | |
| Price | 8.426 | 12.79 | 6.497 | 11.22 | 7.261 | 11.79 | 6.573 | 11.49 | 7.568 | 11.36 |
| Clean | 12.89 | | 11.53 | | 11.98 | | 11.83 | | 11.92 | |
| MPE | 17.06 | | 15.63 | | 16.13 | | 16.07 | | 16.10 | |
| Price | 13.69 | 17.47 | 12.60 | 16.43 | 12.66 | 16.43 | 13.77 | 17.73 | 12.35 | 16.08 |
| std=10% | 18.70 | | 16.61 | | 16.55 | | 17.96 | | 16.18 | |
| MPE | 20.02 | | 20.08 | | 20.10 | | 21.46 | | 19.72 | |
| Price | 29.09 | 36.83 | 29.41 | 36.67 | 29.14 | 36.74 | 31.50 | 38.43 | 29.25 | 36.96 |
| std=30% | 39.99 | | 39.91 | | 39.99 | | 41.80 | | 40.15 | |
| MPE | 41.40 | | 40.69 | | 41.10 | | 41.98 | | 41.47 | |
| Rule (Chaotic) | | 3 | | 2 | | 3 | | 2 | | 3 |
| Rule (power) | | 2 | | 3 | | 3 | | 1 | | 2 |
| Rule (price) | | 2 | | 2 | | 2 | | 2 | | 2 |